\documentclass[11pt]{article}
\usepackage[margin=1in]{geometry}
\usepackage[T1]{fontenc}
\usepackage{times}
\usepackage{graphicx}
\usepackage{booktabs}
\usepackage{multirow}
\usepackage[section]{placeins}
\usepackage{amsmath,amssymb}
\usepackage{xcolor}
\usepackage{natbib}
\usepackage{hyperref}
\hypersetup{colorlinks=true,linkcolor=blue,citecolor=blue,urlcolor=blue}
\usepackage{microtype}
\title{Real Data Closes Synthetic-to-Real\\
Gap in Optical Chemical Structure Recognition}
\author{Yani Guan\thanks{These authors contributed equally to this work.}, Dengpan Dong\footnotemark[1], Zi Wei, Shuang Luo, Dan Hannah\thanks{Corresponding authors: \texttt{dan.hannah@ses.ai}, \texttt{yumin.zhang@ses.ai}, \texttt{kang.xu@ses.ai}},
Yumin Zhang\footnotemark[2],
Kang Xu\footnotemark[2]\\
SES AI Corporation}
\date{}

\begin{document}
\maketitle

\begin{abstract}
Millions of chemical structures appear in patents and papers only as drawings. A system must be able to read these drawings to use this information. OCSR appears nearly solved on synthetic images, but remains difficult on real documents. The starting recognizer Qwen2.5-VL-7B in this work achieves over $91\%$ accuracy on synthetic images, but below $16\%$ on three real-world benchmarks: ACS, CLEF-IP, and USPTO. We fine-tuned 21 recognizers on mixtures of synthetically rendered structures and labeled real depictions drawn from patents, journal figures, and hand-drawn collections, in order to identify the main source of improvement. The experiments varied the Vision Language Model (VLM) base, the fraction of real training data, and the vision-tower adaptation strategy.  
Adding labeled real training images makes the largest difference. For Qwen2.5-VL, ACS exact match rises from $0.15$ with no real data to $0.37$ at $9.5\%$ real data and $0.46$ at $50.2\%$. A controlled experiment across three base models shows the same trend. In contrast, adding a vision LoRA does not improve Qwen at all ($+0.00$, paired $p{=}1.00$). It substantially helps InternVL3-8B ($+22.8$ to $+34.6$\,pt) and modestly helps GLM-4.1V-9B ($+1.0$ to $+9.6$\,pt), showing that its value depends on the base model. It achieves $0.96$ exact match on clean renders and $0.49$, $0.65$, $0.84$, and $0.76$ on ACS, CLEF-IP, UOB, and USPTO, respectively. Differences between base models are largest without real data ($0.21$) and shrink at $70\%$ real data ($0.06$), while their ranking also changes. The base model and real-data mixture must therefore be selected together. Small-scale experiments on handwritten image-to-LaTeX recognition and chart-to-table conversion show that base-model rankings also vary beyond chemistry. More generally, model and adaptation choices for visual structure recognition should be evaluated on the target task.

\end{abstract}

\section{Introduction}
Chemists often communicate structures by drawing 2D graphs because human brains are programmed in decoding such representations better \citep{larkin1987,nieder2025}. However, in computation chemistry, AI and ML models, it is often necessary to convert those drawings into machine-readable structures, such as SMILES, SELFIE or MOL representations, so that literature corpus of patents, journals, books and lab notebooks can be usable for calculation, training, search, property prediction, and synthesis design.

Modern Optical Chemical Structure Recognition (OSCR) systems differ mainly in how much molecular structural data they build into the recognition process. DECIMER \citep{decimer} treats OCSR as direct image-to-SMILES translation, which is simple and scalable but does not explicitly reconstruct the molecular graph. MolScribe \citep{molscribe} builds in a stronger structural prior, detecting atoms and bonds first and assembling the graph from them, although local detection errors can then propagate to the final molecule. MolNexTR \citep{molnextr} further combines a convolutional image encoder with a transformer-based graph decoder to capture both local visual features and long-range connectivity. These specialist models provide useful task-specific constraints and are relatively efficient, but they remain tied to a fixed recognition task and the depiction styles seen during training. Fine-tuned vision--language models (VLMs) offer a more flexible alternative by adapting general visual and language representations to OCSR, often through lightweight methods such as LoRA. 

In this work, an initial Qwen2.5-VL-7B model \citep{qwenvl} fine-tuned with LoRA \citep{lora} achieves $0.95$ exact-match on clean RDKit \citep{rdkit} renders, and
stays above $0.91$ across several synthetic variations e.g. CoordGen re-layout, Indigo cross-toolkit rendering, and synthetic scan/watermark/noise degradation. However, its exact match rapidly falls to $0.15$ on real ACS journal figures
and to approximately $0.12$ on CLEF-IP and USPTO, showing that synthetic performance does not linearly extrapolate to performance on real documents. To close this gap, several data-centric methods are tested, such as larger synthetic datasets, additional rendering toolkits, stronger degradation, reinforcement learning, different VLM bases, and different LoRA targets. The clearest improvement comes from real training images: with the same Qwen base and language-decoder-only LoRA, adding $10\%$ real data raises ACS exact match from $0.15$ to $0.37$, while $50\%$ real data further raises it to $0.46$. In contrast, adding a vision LoRA to the matched synthetic-only Qwen recipe does not cause any change to ACS accuracy. It is also the only other intervention measured on real documents: more synthetic images, more toolkits, heavier degradation, and reinforcement learning were all scored on rendered depictions alone. Our strongest overall model is GLM-4.1V-9B, trained on a mixture of synthetic and real depictions with LoRA applied to the language decoder, vision tower, and connector. It achieves $0.96$ exact match on clean renders and $0.49$, $0.65$, $0.84$, and $0.76$ on ACS, CLEF-IP, UOB, and USPTO, respectively.

\paragraph{Contributions}\mbox{}\par
\noindent

(1) Real labeled training images are the most effective way to close the synthetic-to-real gap. We confirm this with both a controlled three-base, six-fraction experiment (\S\ref{sec:realfrac}) and an independent Qwen dose series (\S\ref{sec:realscale}).

(2) Vision LoRA does not always help. On synthetic-only Qwen2.5-VL-7B, extending LoRA to the vision tower and connector changes ACS exact match by $0.000$ ($p=1.00$).

(3) The value of vision adaptation depends on the base model. It provides large gains on InternVL3-8B ($+22.8$ to $+34.6$\,pt), small gains on GLM-4.1V-9B, and no gain on Qwen2.5-VL-7B (\S\ref{sec:vision}).

(4) Base-model choice matters most when real data are scarce. The performance spread between bases falls from $0.212$ with no real data to $0.060$ at $70\%$ real data, and their ranking changes as more real data are added (\S\ref{sec:realfrac}).

(5) We compare our models with DECIMER, MolScribe, MolNexTR, OCSRGlyph, MarkushGlyph \citep{glyph,decimer2,molscribe,molnextr}, and released chemical VLMs using the same images, labels, and evaluation protocol.
(\S\ref{sec:engines}--\ref{sec:chemvlm}).

\section{Related Work}
\paragraph{OCSR engines.}
The development of OCSR has been more than a shift from rules to larger neural networks; it has also changed how molecular structure is represented during recognition. OSRA \citep{osra} relies on hand-designed rules for detecting graphical elements and is therefore sensitive to changes in drawing style and image quality. DECIMER \citep{decimer} replaces these rules with direct image-to-SMILES generation trained largely on RanDepict \citep{randepict} images, showing that synthetic data can scale recognition but also tying performance closely to the training renderers. MolScribe \citep{molscribe} moves from sequence transcription to atom-and-bond graph reconstruction and includes USPTO depictions, introducing a stronger chemical inductive bias. MolNexTR \citep{molnextr} develops this direction further by combining a ConvNeXt encoder with a transformer graph decoder to capture both local symbols and long-range connectivity. 

\paragraph{VLM-based recognition.} VLMs take a different route: rather than building a dedicated molecular-image decoder, they adapt a general multimodal model such as Qwen2.5-VL \citep{qwenvl}, InternVL3 \citep{internvl}, or GLM-4.1V \citep{glm} to generate SMILES from an image, typically using LoRA\citep{lora}. Their advantage is broad visual pretraining and a flexible text interface, which make a new recognizer relatively inexpensive to fine-tune.

\paragraph{Chemistry-specific VLMs.}
ChemVLM \citep{chemvlm} and TinyChemVL \citep{tinychemvl} adapt general VLMs specifically for chemical tasks, including image-to-SMILES recognition. ChemVLM combines a visual encoder with a chemistry language model, while TinyChemVL uses fewer visual tokens to reduce computation. Both are mainly evaluated using Tanimoto similarity on ChemOCR/img2smiles.

\paragraph{Benchmarks.}
Models are evaluated on held-out images from ACS \citep{molscribe}, CLEF-IP, UOB, USPTO, Staker \citep{staker}, and the hand-drawn DECIMER-HDM dataset \citep{decimerhdm}. Together, these benchmarks cover journal figures, patents, and hand-drawn structures. Because large public training collections of real journal figures remain unavailable, we also build a literature-derived corpus in Appendix~\ref{app:consensus}. Molecular identity is compared using RDKit \citep{rdkit} and InChIKey \citep{inchi}.  

Two small cross-task probes are additionally included: 
image-to-LaTeX transcription \citep{im2latex} and chart-to-table conversion
\citep{chartqa}. Like OCSR, both tasks require an image to be converted into
an exact, machine-readable structure. These probes test whether the observed
base-model dependence extends beyond chemistry, although each configuration
is represented by only one training run.






\section{Experimental Design and the Synthetic-to-Real Gap}
\label{sec:design-gap}

The synthetic-to-real gap is first characterized under a common
training and evaluation framework. Three sources of variation are
separated: the fraction of real training depictions, the pretrained
VLM base, and the adaptation surface. This separation is required
because changes in training data, model architecture, and trainable
modules can otherwise produce the same observed performance difference.

\begin{figure}[t]
    \centering
    \includegraphics[width=\textwidth]{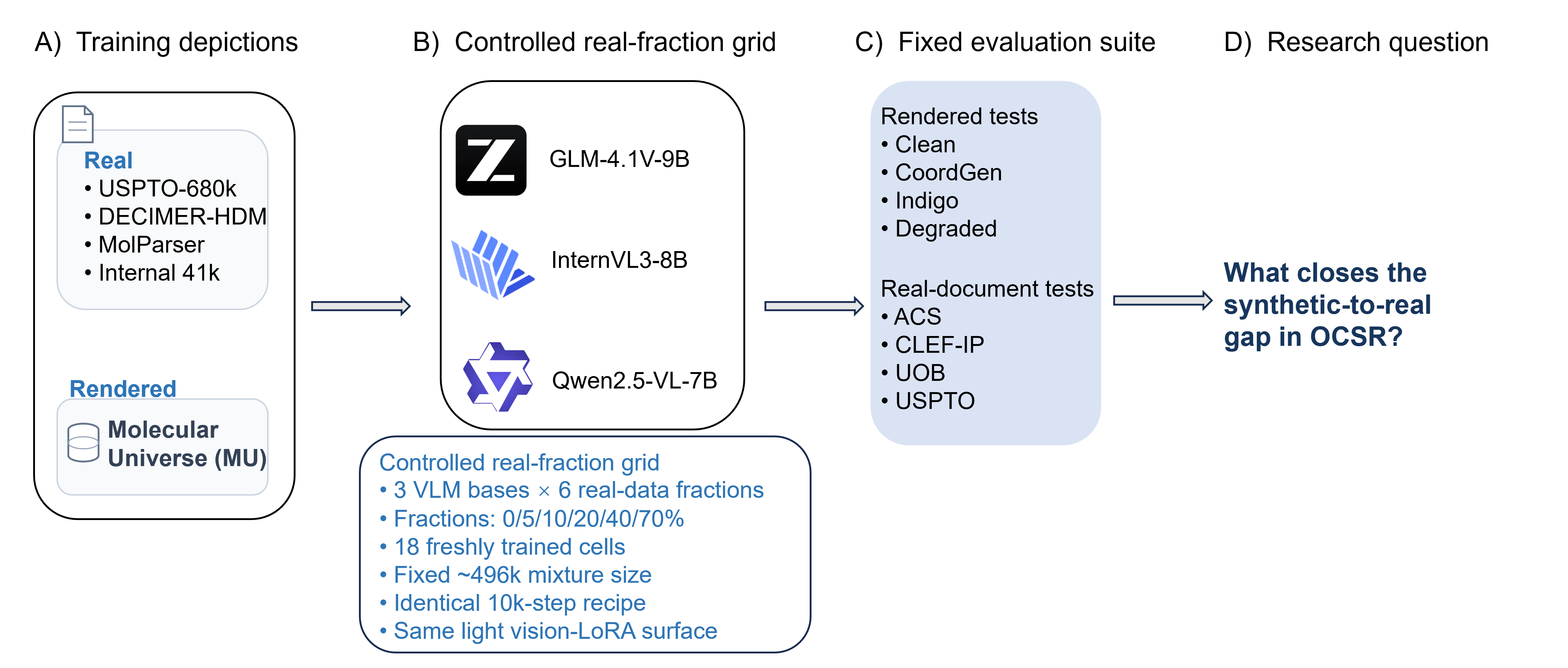}
    \caption{A controlled $3\times6$ design isolates the effects of VLM base
and real-data fraction under a fixed training recipe.}
    \label{fig:overview}
\end{figure}

Figure~\ref{fig:overview} summarizes the study design. The central
controlled experiment crosses three pretrained VLM bases with six
real-data fractions, producing 18 freshly trained cells. The total
mixture size, training budget, light vision-LoRA surface, and evaluation
suite are held fixed across this grid. Additional adaptation-surface
effects are estimated through separate matched contrasts rather than
being treated as part of a complete
$\text{base}\times\text{fraction}\times\text{surface}$ factorial design.

\subsection{Controlled factors}
\label{sec:controlled-factors}

Four sources of variation are distinguished: the pretrained
multimodal stack \(b\), the training-data recipe \(d\), the
optimization objective \(o\), and the adaptation surface \(s\).
For an input image \(x\), the recognizer is written as
\[
f_{b,\phi}(x)=
D_{\theta_{\mathrm{llm}},\phi}\!\left(
C_{\theta_{\mathrm{conn}}}\!\left(
E_{\theta_{\mathrm{vis}}}(x)
\right)
\right),
\]
where \(E\) is the vision encoder, \(C\) is the connector, \(D\) is
the language decoder, and \(\phi\) denotes the downstream adaptation
parameters. The fitted parameters are determined by the data recipe,
objective, and adaptation surface, and are therefore written as
\(\phi(d,o,s)\).

The adaptation surface \(s\) specifies which components are reached
by the downstream parameters. An LM-only surface adapts projections
in the language decoder while leaving the vision encoder and
connector frozen. A vision-adapted surface additionally reaches
selected projections in the vision encoder and connector. This
variable is treated explicitly because runs using the same base and
training images can differ substantially if pretrained representations are allowed to change.

The central controlled experiment crosses three pretrained bases---
Qwen2.5-VL-7B, InternVL3-8B, and GLM-4.1V-9B---with six real-data
fractions:
\[
0\%,\ 5\%,\ 10\%,\ 20\%,\ 40\%,\ 70\%.
\]
The resulting \(18\) cells are freshly fine-tuned from their
respective pretrained bases. An approximately \(496\)k-sample
mixture size, a \(10\)k-step budget, the supervised fine-tuning
objective, and the light vision-LoRA surface are held fixed across
the grid. Increasing the real-data fraction therefore replaces
rendered examples rather than increasing the total number of
training examples.

Within this grid, the real-data effect is estimated at fixed base,
objective, and adaptation surface, while base effects are compared
at fixed real-data fractions under the same recipe. Adaptation-surface effects are estimated separately, through matched within-base contrasts: LM-only versus vision-plus-connector LoRA on synthetic-only Qwen2.5-VL-7B, and
frozen versus light-LoRA vision towers on GLM-4.1V-9B and InternVL3-8B
trained with real depictions. 

These paiors are not cells of a full
$\text{base}\times\text{real fraction}\times\text{adaptation surface}$
design. The reduction is deliberate rather than imposed: every base
admits every surface under the same training configuration, so the
complete design is realizable, but at $54$ cells it would require
approximately twice the total training cost reported in
Table~\ref{tab:compute}. Effort was therefore concentrated on the two
factors carrying the principal claims, with the third sampled through
matched contrasts. One level of that factor is in any case
unrecoverable: the fully unfrozen arms admit no evaluation, their
trained vision weights having never been persisted
(\S\ref{sec:vision}).

\subsection{Fixed evaluation suite}
\label{sec:evaluation-suite}

All controlled configurations are evaluated using the same image
sets, prompts, generation settings, canonicalization procedure, and
scoring implementation. The rendered suite contains clean RDKit
depictions, CoordGen relayouts, Indigo cross-toolkit renders, and
synthetically degraded images. The real-document suite contains ACS
journal figures, CLEF-IP patent figures, UOB hand-drawn structures,
and USPTO patent figures.

Each rendered condition contains up to \(500\) held-out molecules,
with \(369\) images available for the Indigo condition. The ACS set
contains \(331\) images, while CLEF-IP, UOB, and USPTO are evaluated
with a common cap of \(500\) images. Identical evaluation images are
used for every configuration within a comparison, allowing
per-image paired tests to be applied.

The primary metric is canonical exact match with full
stereochemistry. A generated SMILES is first parsed and canonicalized
with RDKit, and molecular identity is determined using the
corresponding InChIKey. An invalid or nonmatching prediction receives
an exact-match score of zero. For an evaluation set
\(\mathcal{D}=\{(x_i,y_i)\}_{i=1}^{n}\), accuracy is defined as
\[
A(\mathcal{D})=
\frac{1}{n}
\sum_{i=1}^{n}
\mathbf{1}
\left\{
\operatorname{key}(\hat{y}_i)
=
\operatorname{key}(y_i)
\right\},
\]
where \(\operatorname{key}(\cdot)\) denotes the canonical molecular
identity obtained after RDKit parsing. Validity is reported as a
secondary diagnostic:
\[
V(\mathcal{D})=
\frac{1}{n}
\sum_{i=1}^{n}
\mathbf{1}
\left\{
\operatorname{parse}(\hat{y}_i)
\text{ succeeds}
\right\}.
\]
Validity distinguishes failure to produce a chemically interpretable
output from production of a valid but incorrect molecular structure.

The real-document sets represent different degrees of domain
transfer. ACS provides the clearest held-out journal-depiction test.
UOB and USPTO are closer in style to hand-drawn and patent sources
represented in the real training pool. Molecular-identity
deduplication removes overlapping structures, but it does not remove
similarities in depiction style or document source. ACS is therefore
used as the primary held-out-domain result, while all four real sets
are reported to expose condition-specific behavior.

The degraded rendered condition should also be interpreted
separately from real-document robustness. Its corruption process is
synthetically constructed and is represented in the training recipe;
performance on this condition therefore measures robustness to the
specified corruption model rather than transfer to previously unseen
real scans.


\subsection{Strong Rendered Accuracy Does Not Transfer to Real Documents}
\label{sec:gap}

A synthetic-only Qwen2.5-VL-7B checkpoint is used to establish the
initial domain gap. The model is adapted with LM-only LoRA, while the
vision encoder and connector remain frozen. Canonical exact match
reaches \(0.950\) on clean RDKit depictions, \(0.938\) on CoordGen
relayouts, \(0.957\) on Indigo cross-toolkit renders, and \(0.914\)
after synthetic degradation. The average exact match across these
four rendered conditions is \(0.940\).

Performance is substantially lower on real-document images. Exact
match falls to \(0.154\) on ACS, \(0.124\) on CLEF-IP, \(0.522\) on
UOB, and \(0.122\) on USPTO, corresponding to a four-set average of
\(0.231\). UOB is a clear exception to the otherwise uniformly low
real-document scores, but the available experiment does not identify
which depiction characteristics account for this difference.

The decline in exact match is not explained solely by invalid SMILES.
Validity is \(0.719\) on ACS, \(0.724\) on CLEF-IP, \(0.896\) on UOB,
and \(0.730\) on USPTO. Thus, chemically parseable outputs are
frequently produced even when the predicted molecular identity is
incorrect. High accuracy on rendered depictions therefore does not
establish reliable recognition of molecular structures in real
scientific documents.

This baseline establishes the problem but does not identify its
cause. Synthetic scale, renderer diversity, degradation augmentation,
optimization objective, real-data fraction, pretrained base, and
adaptation surface are examined in the following sections.
Section~\ref{sec:real-data} first compares synthetic interventions
with the addition of representative real training images.

\section{Representative Real Data Closes the Gap}
\label{sec:real-data}

Interventions are separated according to whether they improve only
the rendered condition they simulate or transfer consistently to
real scientific documents. Synthetic scale, renderer diversity,
degradation augmentation, and a verifiable reinforcement-learning
objective are first evaluated as lower-cost alternatives to collecting
labeled real depictions. Their effects are compared with those obtained
by replacing rendered training examples with representative real
figures.

Evidence for the real-data effect is presented at two levels. An
independent Qwen2.5-VL-7B dose series first shows that large
real-document gains can be obtained while the vision encoder and
connector remain frozen. This already locates the limitation: not in
the visual representation, which is left unchanged, but in the
supervision that maps it onto correct structures. Because the runs in
this series differ in mixture composition and training budget, the
series is treated as descriptive evidence. The effect is then tested in a controlled three-base grid containing six real-data fractions and 18 freshly
fine-tuned cells. Together, these experiments distinguish improvements
caused by representative real supervision from improvements restricted
to matched rendered conditions.

\subsection{Synthetic Scaling and Recipe Changes Remain Condition-Specific}
\label{sec:synthetic-interventions}

Increasing the number of synthetic training molecules from \(200\)k
to \(7\)M produces no monotonic improvement on the three available
evaluation splits. IID exact match remains within
\(0.900\)--\(0.914\), scaffold-disjoint exact match within
\(0.894\)--\(0.918\), and external PubChem exact match within
\(0.300\)--\(0.358\). The largest variation is observed on external
PubChem depictions, but it is not associated monotonically with the
number of training molecules.

A similar pattern is obtained when the number of rendered training
images is increased from \(100\)k to \(970\)k: clean exact match
remains within \(0.908\)--\(0.934\) and augmented exact match within
\(0.702\)--\(0.738\). Each molecule is rendered once in these runs, so
image count and molecule count grow together and the two synthetic
axes are not separated. Under the multi-renderer recipe, which renders
each molecule twice, the augmented split improves modestly with image
count, but clean accuracy stays well below the single-renderer
results. Additional synthetic scale therefore does not provide a
consistent improvement across the rendered evaluation conditions.

\begin{table}[!htbp]
\centering
\small
\caption{
Increasing synthetic-data scale does not consistently improve
accuracy on rendered images.
}
\label{tab:scaling}
\begin{tabular}{lccc}
\toprule
Training molecules
    & IID
    & Scaffold OOD
    & External PubChem \\
\midrule
\(200\)k & .900 & .916 & .342 \\
\(500\)k & .904 & .918 & .358 \\
\(1\)M   & .902 & .910 & .302 \\
\(2\)M   & .908 & .910 & .352 \\
\(3\)M   & .906 & .894 & .342 \\
\(5\)M   & .914 & .918 & .310 \\
\(7\)M   & .910 & .910 & .300 \\
\midrule
Rendered images
    & Clean
    & Augmented
    & \\
\midrule
\(100\)k & .930 & .716 & \\
\(200\)k & .932 & .716 & \\
\(300\)k & .926 & .706 & \\
\(400\)k & .934 & .722 & \\
\(500\)k & .910 & .738 & \\
\(600\)k & .922 & .730 & \\
\(700\)k & .908 & .702 & \\
\(800\)k & .916 & .706 & \\
\(900\)k & .916 & .720 & \\
Maximum available & .922 & .722 & \\
\midrule
\multicolumn{4}{l}{\emph{Multi-renderer mixture}} \\
\(200\)k & .716 & .546 & \\
\(500\)k & .716 & .580 & \\
\(900\)k & .706 & .600 & \\
\bottomrule
\end{tabular}
\end{table}

Additional rendered-only interventions produce similarly limited
changes. A verifiable RDKit reward increases clean exact match from
\(0.930\) to \(0.944\), while augmented, CoordGen, Indigo, and degraded
performance remain approximately unchanged. Thus, the optimization
objective provides a small gain on the condition that is already
strongest but does not produce a consistent robustness improvement.

At a matched \(200\)k-image scale, extending LoRA to the vision tower
reduces clean exact match by \(2.2\) points and CoordGen exact match by
\(7.0\) points. Fully unfreezing the vision tower produces larger
declines on the same rendered conditions. These results indicate that
increasing the trainable vision capacity does not automatically improve
rendered recognition and can disturb useful pretrained visual features.

The multi-renderer result is also unfavorable at a matched image
count. At \(500\)k images, the single-renderer recipe reaches \(0.910\)
on clean and \(0.738\) on augmented depictions
(Table~\ref{tab:scaling}), whereas the multi-renderer recipe reaches
only \(0.716\) and \(0.580\), respectively. Renderer diversity therefore
does not compensate for the loss in clean rendered accuracy in these
experiments.

\begin{table}[!htbp]
\centering
\small
\caption{
Rendered-only interventions provide small or negative gains on
Qwen2.5-VL.
}
\label{tab:interventions}
\setlength{\tabcolsep}{5pt}
\begin{tabular}{lccccc}
\toprule
Configuration
    & Clean
    & Augmented
    & CoordGen
    & Indigo
    & Degraded \\
\midrule
SFT v2 baseline
    & .930 & .752 & .938 & .940 & .502 \\
\(+\) RL with verifiable RDKit reward (GRPO)
    & .944 & .748 & .936 & .935 & .508 \\
Single-renderer, \(200\)k images
    & .932 & .716 & .720 & .420 & .430 \\
\(+\) LoRA on the vision tower
    & .910 & .692 & .650 & --- & .388 \\
\(+\) vision tower fully unfrozen
    & .858 & .656 & .640 & --- & .350 \\
Multi-renderer, \(500\)k images
    & .716 & .580 & .716 & --- & .332 \\
\bottomrule
\end{tabular}
\end{table}

Because the interventions in Table~\ref{tab:interventions} were
evaluated only on rendered conditions, no conclusion about
real-document transfer can be obtained from these runs alone.
They show that additional synthetic scale, a verifiable reward,
renderer diversity, and a larger trainable vision surface provide
either small, inconsistent, or negative changes on the available
rendered evaluations. The effect of representative real supervision
is examined separately in Sections~\ref{sec:realscale}
and~\ref{sec:realfrac}.

\subsection{An Independent Qwen Real-Data Dose Series}
\label{sec:realscale}

An independent dose series is assembled from six Qwen2.5-VL-7B
checkpoints whose training mixtures contain between \(0\%\) and
\(50.2\%\) labeled real depictions. The same LM-only LoRA target set
is used throughout, while the vision encoder and connector remain
frozen. Consequently, any improvement observed in this series does
not require adaptation of the vision path.

The series was not designed as a controlled fraction sweep. Training
mixture composition, step budget, and other recipe details vary across
the checkpoints. The results are therefore treated as descriptive
evidence for the effect of adding real images rather than as an
estimate of a smooth dose-response curve or an optimal real-data
fraction.

\begin{table}[!htbp]
\centering
\small
\caption{
More real training data generally improves real-document accuracy under LM-only LoRA.
}
\label{tab:realdose}
\setlength{\tabcolsep}{5pt}
\begin{tabular}{@{}lccccc@{}}
\toprule
Run
    & Real fraction
    & ACS
    & CLEF-IP
    & UOB
    & USPTO \\
\midrule
\texttt{sft\_v4}
    & \(0\%\)
    & .154/.719
    & .124/.724
    & .522/.896
    & .122/.730 \\
\texttt{sft\_v5}
    & \(0\%\)
    & .169/.689
    & .140/.678
    & .518/.878
    & .116/.716 \\
\texttt{sft\_real}
    & \(9.5\%\)
    & .372/.816
    & .608/.888
    & .818/.986
    & .698/.924 \\
\texttt{sft\_v6}
    & \(11.5\%\)
    & .347/.852
    & .604/.890
    & .808/.976
    & .742/.936 \\
\texttt{sft\_v5\_morereal}
    & \(41.6\%\)
    & .402/.825
    & .628/.876
    & .842/.992
    & .824/.954 \\
\texttt{sft\_qwen\_vr1}
    & \(50.2\%\)
    & .459/.888
    & .614/.918
    & .832/.988
    & .728/.968 \\
\bottomrule
\end{tabular}
\end{table}

The two synthetic-only checkpoints establish a narrow reference range:
ACS exact match is \(0.154\)--\(0.169\), CLEF-IP is
\(0.124\)--\(0.140\), UOB is \(0.518\)--\(0.522\), and USPTO is
\(0.116\)--\(0.122\). When \(9.5\%\) real data is introduced, exact
match reaches \(0.372\) on ACS, \(0.608\) on CLEF-IP, \(0.818\) on
UOB, and \(0.698\) on USPTO. These increases are substantially larger
than the changes obtained from the rendered-only interventions in
Section~\ref{sec:synthetic-interventions}.

The held-out ACS result is especially informative because it is less
closely related to the real training sources than the patent and
hand-drawn evaluations. ACS exact match increases further to \(0.459\)
at \(50.2\%\) real data. Large gains are also observed on CLEF-IP,
UOB, and USPTO, although these datasets are closer in depiction style
to sources represented in the real training pool.

Validity generally increases after real depictions are introduced.
For example, ACS validity rises from \(0.719\) for
\texttt{sft\_v4} to \(0.816\) at \(9.5\%\) real data and \(0.888\)
at \(50.2\%\). The improvement therefore reflects both a greater
probability of producing a chemically parseable output and a greater
probability of recovering the correct molecular identity.

The point estimates are not monotonic at every fraction or on every
dataset. USPTO reaches \(0.824\) exact match at \(41.6\%\) real data
but falls to \(0.728\) at \(50.2\%\), while smaller reversals are
observed on ACS, CLEF-IP, and UOB. Because fraction, composition, and
training budget change together in this series, these reversals cannot
be attributed to the real-data fraction alone.

The series nevertheless establishes a narrower result: large
real-document gains can be obtained while the vision path remains
frozen. A controlled three-base experiment is presented next to test
whether the real-data effect remains when mixture size, training
budget, objective, and adaptation surface are held fixed.

\subsection{A Controlled Sweep Confirms the Real-Data Effect}
\label{sec:realfrac}

The effect of the real-data fraction is isolated in a controlled
\(3\times6\) sweep. Three pretrained bases are crossed with six
real-data fractions, producing \(18\) freshly fine-tuned cells.
An approximately \(496\)k-sample mixture size, a \(10\)k-step
training budget, the supervised fine-tuning objective, and the light
vision-LoRA adaptation surface are held fixed. Increasing the
real-data fraction therefore replaces rendered training examples
rather than increasing the total number of examples.

\begin{figure}[t]
    \centering
    \includegraphics[width=\textwidth]
    {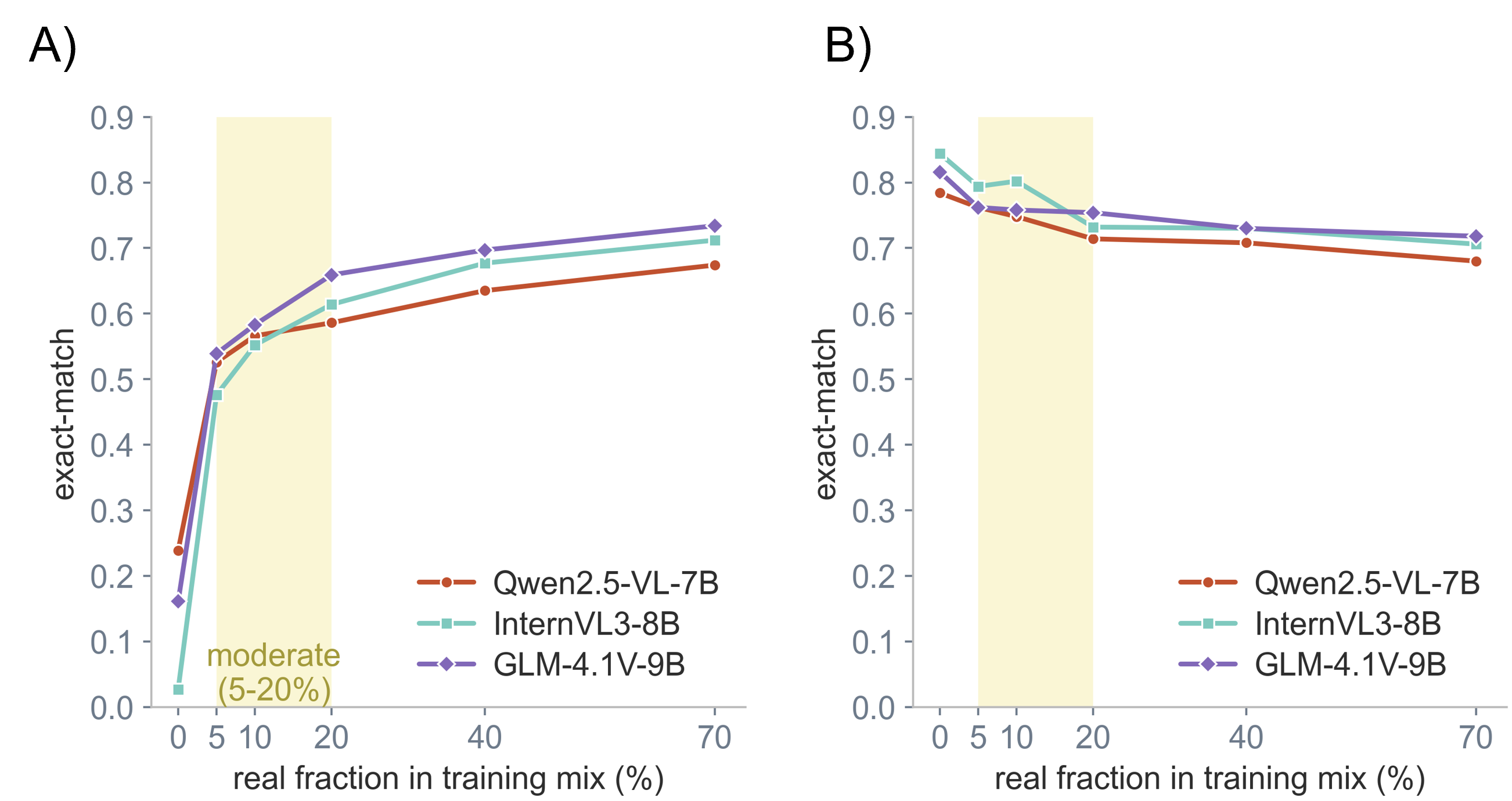}
    \caption{
    More real training images improve real-document accuracy but reduce rendered-clean accuracy.
    }
    \label{fig:realfrac}
\end{figure}

Figure~\ref{fig:realfrac} shows a consistent dependence across all
three bases. Mean real-document exact match increases sharply between
\(0\%\) and \(5\%\) real data and continues to increase through
\(70\%\). The first increment is the largest: Qwen2.5-VL-7B increases
from \(0.239\) to \(0.526\), InternVL3-8B from \(0.027\) to \(0.476\),
and GLM-4.1V-9B from \(0.161\) to \(0.539\). Smaller positive gains
are obtained at every subsequent fraction, and no plateau is observed
within the tested range.

\begin{table}[!htbp]
\centering
\small
\caption{Real-document accuracy increases with the fraction of real training data across all three VLM bases.}
\label{tab:realfrac}
\setlength{\tabcolsep}{5pt}
\begin{tabular}{lcccccc}
\toprule
& \multicolumn{2}{c}{Qwen2.5-VL-7B}
& \multicolumn{2}{c}{InternVL3-8B}
& \multicolumn{2}{c}{GLM-4.1V-9B} \\
\cmidrule(lr){2-3}
\cmidrule(lr){4-5}
\cmidrule(lr){6-7}
Real fraction
    & Real avg.
    & Clean
    & Real avg.
    & Clean
    & Real avg.
    & Clean \\
\midrule
\(0\%\)  & .239 & .784 & .027 & .844 & .161 & .816 \\
\(5\%\)  & .526 & .762 & .476 & .794 & .539 & .762 \\
\(10\%\) & .566 & .748 & .552 & .802 & .583 & .758 \\
\(20\%\) & .586 & .714 & .614 & .732 & .659 & .754 \\
\(40\%\) & .635 & .708 & .677 & .730 & .697 & .730 \\
\(70\%\) & .674 & .680 & .712 & .706 & .734 & .718 \\
\bottomrule
\end{tabular}
\end{table}

The initial \(5\%\) of real data accounts for approximately two-thirds
of the total real-document gain observed between \(0\%\) and \(70\%\)
real data for every base. By \(20\%\) real data, approximately
\(79.8\%\) of the total Qwen gain, \(85.7\%\) of the total InternVL
gain, and \(86.9\%\) of the total GLM gain have already been obtained.
The shaded \(5\%\)--\(20\%\) interval therefore represents a useful
moderate-data regime, although it is not established as an optimum.

The real-document improvement is accompanied by a decline in
rendered-clean exact match under the fixed-size mixture. Between
\(0\%\) and \(70\%\) real data, the rendered-clean score decreases
from \(0.784\) to \(0.680\) for Qwen, from \(0.844\) to \(0.706\)
for InternVL, and from \(0.816\) to \(0.718\) for GLM. These changes
correspond to losses of \(10.4\), \(13.8\), and \(9.8\) percentage
points, respectively. The real-data fraction must therefore be
selected as an operating point between real-document transfer and
rendered-domain retention.

The base ordering also changes across the sweep. Qwen has the highest
real-document average at \(0\%\) real data, while GLM leads at every
nonzero fraction. InternVL moves from the weakest base at \(0\%\) to
the second strongest from \(20\%\) onward. The interaction between
base choice and real-data fraction is examined in
Section~\ref{sec:model-adaptation}.

\section{Model and Adaptation Choices Depend on the Real-Data Regime}
\label{sec:model-adaptation}

The controlled real-fraction sweep establishes that representative
real supervision improves real-document recognition across all three
pretrained bases. This shared dependence does not eliminate the effects
of model architecture or adaptation strategy. Instead, the magnitude
and ordering of those effects change with the amount of real
supervision available.

Three questions are addressed in this section. First, base-model
rankings are compared across the controlled real-fraction grid to
determine whether a ranking measured under synthetic-only training
remains valid after real depictions are introduced. Second,
adaptation-surface effects are examined through matched within-base
contrasts, with the distinction maintained between synthetic-only
Qwen and real-data GLM and InternVL experiments. Finally, representative
operating points are compared to show how rendered accuracy,
real-document accuracy, and adaptation strategy jointly affect model
selection.

\subsection{Base Rankings Change with the Real-Data Budget}
\label{sec:base}

Base effects are first examined at \(0\%\) real data, where the
training mixture, step budget, optimization objective, and adaptation
surface are matched across all three pretrained stacks. Under this
synthetic-only condition, Qwen2.5-VL-7B obtains the highest mean
real-document exact match at \(0.239\), followed by GLM-4.1V-9B at
\(0.161\) and InternVL3-8B at \(0.027\). The resulting between-base
spread is \(0.212\), substantially larger than the spread observed at
any nonzero real-data fraction.

The Qwen advantage at \(0\%\) real data is not uniform across all
datasets. Qwen leads on CLEF-IP, UOB, and USPTO, while GLM is slightly
higher on ACS (\(0.139\) versus \(0.121\)). InternVL is substantially
lower on all four real-document sets despite obtaining the highest
rendered-clean score. Thus, rendered-clean accuracy does not determine
which pretrained base transfers most effectively to real documents
under synthetic-only fine-tuning.

\begin{table}[!htbp]
\centering
\small
\caption{
VLM base choice strongly affects transfer from rendered training to real documents.
}
\label{tab:base}
\resizebox{\textwidth}{!}{%
\begin{tabular}{lccccc}
\toprule
Base
    & Rendered clean
    & ACS
    & CLEF-IP
    & UOB
    & USPTO \\
\midrule
Qwen2.5-VL-7B
    & 0.784
    & 0.121/0.671
    & 0.162/0.774
    & 0.528/0.908
    & 0.146/0.718 \\
InternVL3-8B
    & 0.844
    & 0.024/0.486
    & 0.006/0.516
    & 0.068/0.674
    & 0.010/0.516 \\
GLM-4.1V-9B
    & 0.816
    & 0.139/0.565
    & 0.056/0.440
    & 0.404/0.824
    & 0.044/0.504 \\
\bottomrule
\end{tabular}%
}
\end{table}

The ranking changes after real depictions are introduced. At
\(5\%\) and \(10\%\) real data, the ordering becomes GLM, Qwen, and
InternVL. From \(20\%\) through \(70\%\), the ordering becomes GLM,
InternVL, and Qwen. InternVL therefore moves from the weakest
synthetic-only base to the second strongest base under moderate and
high real-data fractions.

The separation between bases also contracts sharply. The spread is
\(0.212\) at \(0\%\) real data, falls to \(0.063\) at \(5\%\), and
remains between \(0.031\) and \(0.073\) across all nonzero fractions.
At \(70\%\) real data, the spread is \(0.060\). Base choice therefore
has its largest observed effect when representative real supervision
is absent, while differences between pretrained stacks become much
smaller after real depictions are added.

These comparisons characterize the complete pretrained multimodal
stack rather than an isolated architectural component or a parameter
scaling law. A change of base simultaneously changes the vision
encoder, connector, language decoder, image processor, native
resolution, visual tokenization, chat template, and unknown
pretraining exposure. The controlled sweep establishes that the base
and real-data fraction must be selected jointly, but it does not
identify which component of the pretrained stack causes the observed
ranking.

\subsection{Vision Adaptation Is Base- and Data-Dependent}
\label{sec:vision}
\label{sec:surface}

Adaptation-surface effects are estimated through three matched
within-base contrasts. The contrasts are not treated as cells of a
complete
\[
\text{base}\times\text{real fraction}\times\text{adaptation surface}
\]
factorial design. In particular, the Qwen experiment uses synthetic-only
training data and compares LM-only LoRA with LoRA extended jointly to
the vision encoder and connector. The GLM and InternVL experiments
include real depictions and compare a frozen vision tower with light
vision LoRA; the connector is adapted in both arms. Effect sizes are
therefore interpreted within each panel rather than averaged across
the three bases.

\paragraph{Synthetic-only Qwen.}
For Qwen2.5-VL-7B, the data mixture, \(18\)k-step budget, LoRA rank,
LoRA alpha, learning rate, and pretrained base are held fixed.
The only change is the extension of the LoRA target set from seven
language-decoder module types to additional projections in the vision
encoder and connector.

\begin{table}[!htbp]
\centering
\small
\caption{
Vision LoRA does not improve synthetic-only Qwen2.5-VL-7B.
}
\label{tab:surface}
\setlength{\tabcolsep}{5pt}
\begin{tabular}{lccccc}
\toprule
LoRA target set
    & Rendered clean
    & ACS
    & CLEF-IP
    & UOB
    & USPTO \\
\midrule
LM only (\(7\) module types)
    & .950/.998
    & .1541/.719
    & .124/.724
    & .522/.896
    & .122/.730 \\
Vision \(+\) connector (\(10\) module types)
    & .956/1.000
    & .1541/.710
    & .102/.712
    & .486/.868
    & .122/.724 \\
\midrule
\(b/c\)
    & ---
    & \(11/11\)
    & \(16/27\)
    & \(25/43\)
    & \(22/22\) \\
Paired \(p\)
    & ---
    & \(1.00\)
    & \(0.13\)
    & \(0.038\)
    & \(1.00\) \\
\bottomrule
\end{tabular}
\end{table}

No positive real-document effect is obtained from extending the Qwen
adaptation surface. ACS and USPTO exact match are unchanged, with
perfectly symmetric paired disagreements. CLEF-IP decreases from
\(0.124\) to \(0.102\), and UOB decreases from \(0.522\) to \(0.486\).
The UOB decrease is nominally significant at \(p=0.038\), but it does
not survive correction across the four real-document comparisons.
Rendered-clean exact match increases slightly from \(0.950\) to
\(0.956\). Thus, vision-plus-connector LoRA does not close the
real-document gap for synthetic-only Qwen.

\begin{figure}[t]
    \centering
    \includegraphics[width=\textwidth]
    {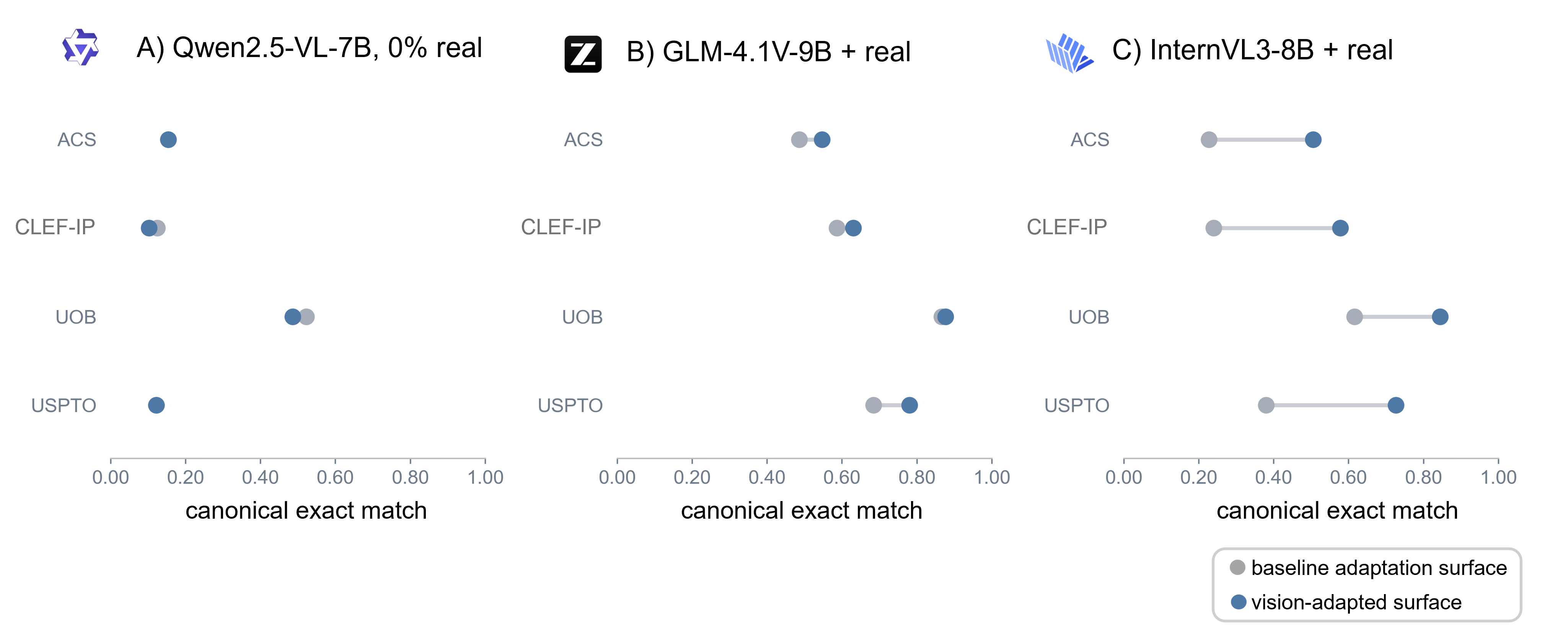}
    \caption{
    The benefit of vision adaptation depends strongly on the VLM base.
    }
    \label{fig:vision}
\end{figure}

\paragraph{GLM and InternVL with real depictions.}
Different behavior is observed when real depictions are present.
Within each base, the frozen and light-LoRA arms use identical training
images and matched training budgets. The resulting paired comparisons
are reported in Table~\ref{tab:vision}.

On GLM-4.1V-9B, light vision LoRA increases exact match from
\(0.486\) to \(0.547\) on ACS, from \(0.586\) to \(0.630\) on
CLEF-IP, from \(0.866\) to \(0.876\) on UOB, and from \(0.684\) to
\(0.780\) on USPTO. The corresponding gains are \(6.1\), \(4.4\),
\(1.0\), and \(9.6\) percentage points. Paired improvements are
detected on ACS and USPTO, while UOB is statistically tied.
The CLEF-IP result has \(p=0.021\) and does not survive a Bonferroni
correction across the four real-document comparisons.

On InternVL3-8B, the matched comparison at step \(9{,}000\) is much
larger. Exact match increases from \(0.227\) to \(0.505\) on ACS,
from \(0.240\) to \(0.578\) on CLEF-IP, from \(0.616\) to \(0.844\)
on UOB, and from \(0.380\) to \(0.726\) on USPTO. These changes
correspond to gains of \(27.8\), \(33.8\), \(22.8\), and \(34.6\)
percentage points, respectively. Every real-set comparison remains
significant beyond \(p<10^{-20}\).

\begin{table}[!htbp]
\centering
\small
\caption{
Light vision LoRA strongly improves InternVL3-8B but provides smaller gains for GLM-4.1V-9B.
}
\label{tab:vision}
\setlength{\tabcolsep}{4.5pt}
\begin{tabular}{llccccc}
\toprule
Base
    & Vision mode
    & Clean
    & ACS
    & CLEF-IP
    & UOB
    & USPTO \\
\midrule
\multirow{3}{*}{GLM-4.1V-9B}
    & Frozen
    & .696
    & .486
    & .586
    & .866
    & .684 \\
    & Light LoRA
    & .726
    & .547
    & .630
    & .876
    & .780 \\
    & Paired \(p\)
    & ---
    & .003
    & .021
    & .33
    & \(<10^{-3}\) \\
\midrule
\multirow{3}{*}{InternVL3-8B}
    & Frozen
    & .730/.984
    & .227/.740
    & .240/.684
    & .616/.914
    & .380/.814 \\
    & Light LoRA
    & .702/.994
    & .505/.922
    & .578/.926
    & .844/.992
    & .726/.964 \\
    & Paired \(p\)
    & .25
    & \(2\!\cdot\!10^{-21}\)
    & \(2\!\cdot\!10^{-43}\)
    & \(4\!\cdot\!10^{-31}\)
    & \(1\!\cdot\!10^{-37}\) \\
\midrule
Either
    & Full unfreeze
    & \multicolumn{5}{c}{
        Not evaluable: trained vision weights were not persisted
      } \\
\bottomrule
\end{tabular}
\end{table}

The InternVL exact-match gains are accompanied by substantial validity
increases. Validity rises from \(0.740\) to \(0.922\) on ACS, from
\(0.684\) to \(0.926\) on CLEF-IP, from \(0.914\) to \(0.992\) on
UOB, and from \(0.814\) to \(0.964\) on USPTO. The effect therefore
includes an increased probability of producing a chemically parseable
output, rather than only improved molecular identity among already
valid predictions.

A small rendered-domain change is observed for InternVL:
rendered-clean exact match decreases from \(0.730\) to \(0.702\).
The paired test does not establish this difference as significant
(\(p=0.25\)); \(56\) images favor light LoRA and \(70\) favor the
frozen tower. The real-document gains are therefore much larger and
more statistically stable than the observed rendered-clean difference.

The full-unfreeze arms cannot be interpreted. Although the runs were
launched with trainable vision weights, the checkpoint path persisted
only adapter tensors, leaving the trained vision-tower weights
unavailable at evaluation time. Scores obtained by loading those
checkpoints would measure a mismatch between a pristine tower and
co-adapted downstream parameters rather than the fully unfrozen model.
No performance claim is therefore made for full vision-tower
unfreezing.

Taken together, the matched contrasts show that vision adaptation is
not a universally beneficial intervention. It is unnecessary for the
synthetic-only Qwen configuration, provides modest and
dataset-dependent gains for GLM with real data, and produces large
gains for InternVL with real data. Because the Qwen and real-data
contrasts are conducted under different data regimes, the observed
differences cannot be attributed to the pretrained base alone. A
complete base-by-data-by-surface experiment would be required to
separate those interactions.

\subsection{Selecting an operating point}
\label{sec:headline}

No single configuration is optimal for every deployment regime. The preferred
operating point depends on whether performance is prioritized on rendered
depictions, held-out journal figures, or document sources closer to the real
training pool. Four representative checkpoints are compared in
Table~\ref{tab:flagship}. These checkpoints were produced by different
training configurations and are therefore presented as deployment-oriented
operating points rather than as a matched causal comparison.

Trained under the same full recipe but on a $70\%$ real mixture, GLM-4.1V-9B
gives the strongest real-document result in this study: ACS $0.598$, CLEF-IP
$0.774$, UOB $0.898$, and USPTO $0.848$, a real-document mean of $0.780$. Its
rendered clean score is $0.734$, and the three remaining rendered conditions
were not evaluated for this checkpoint. It falls below UOB and USPTO because
this checkpoint trades rendered accuracy for real-document accuracy---the same
base reaches $0.962$ on clean at a low real fraction---and because those two
sets are near-in-domain for the real pool; ACS, the only held-out depiction
domain, does fall below it. Relative to the corresponding sweep
cell, which shares the mixture but uses half the step budget, all five measured
conditions improved, rendered clean included ($0.718$ to $0.734$). The
additional training is therefore not paid for out of the rendered-versus-real
trade-off of Section~\ref{sec:realfrac}.

The tuned GLM checkpoint provides the most balanced result among the four
rows. Its mean exact match is $0.952$ across the four rendered tests and
$0.683$ across the four real-document tests. It exceeds the selected InternVL
checkpoint on seven of the eight individual conditions, with synthetic
degradation as the only exception. The InternVL checkpoint nevertheless
retains a similar rendered average of $0.952$ and is the checkpoint used in
the chemical-VLM comparison in Section~\ref{sec:chemvlm}.

A different trade-off is obtained with the light-vision-LoRA GLM checkpoint.
Relative to the tuned GLM checkpoint, its average rendered exact match
decreases from $0.952$ to $0.729$, a loss of $22.4$ percentage points, while
its mean real-document exact match increases from $0.683$ to $0.708$. It is
the better of the two tuned-recipe checkpoints on ACS, UOB, and USPTO, whereas
the tuned GLM checkpoint remains higher on CLEF-IP; both are below the
$70\%$-real checkpoint on all four real sets. The apparent advantage on UOB and USPTO should be interpreted
together with their closer relationship to sources represented in the real
training pool; ACS remains the primary held-out journal domain.

\begin{table}[!htbp]
\centering
\small
\caption{The $70\%$-real GLM checkpoint is strongest on real documents; the tuned GLM checkpoint provides the best balance across rendered and real-document tests.}
\label{tab:flagship}
\setlength{\tabcolsep}{3.5pt}
\resizebox{\textwidth}{!}{%
\begin{tabular}{lcccccccc}
\toprule
configuration
& clean & CoordGen & Indigo & degraded
& ACS & CLEF-IP & UOB & USPTO \\
\midrule
GLM-4.1V-9B $+$ $70\%$ real, full recipe
& .734 & --- & --- & ---
& .598 & .774 & .898 & .848 \\

GLM-4.1V-9B $+$ real, tuned
& .962 & .948 & .959 & .940
& .489 & .650 & .838 & .756 \\

InternVL3-8B $+$ $\sim\!10\%$ real, tuned
& .958 & .938 & .954 & .956
& .459 & .610 & .830 & .732 \\

GLM-4.1V-9B $+$ real, light vision LoRA
& .726 & .734 & .767 & .688
& .547 & .630 & .876 & .780 \\
\bottomrule
\end{tabular}}
\end{table}

These results support operating-point selection rather than a universal model
ranking. High rendered robustness favors the tuned GLM checkpoint, while
greater weight on ACS and the near-domain real-document sets favors the
light-vision-LoRA GLM checkpoint. Comparisons with released specialist OCSR
systems and chemical VLMs are reported in Section~\ref{sec:published} using
the same evaluation images and scoring pipeline.

\section{Comparison with Published Recognizers}
\label{sec:published}

The controlled experiments identify which interventions improve a fine-tuned
VLM, but they do not establish how the resulting checkpoints compare with
existing OCSR systems. Released specialist recognizers and published chemical
VLMs are therefore evaluated using the same test images, label files, image
caps, molecular canonicalization procedure, and exact-match scorer. This
common evaluation avoids comparisons across incompatible test subsets or
stereochemistry conventions.

\begin{figure}[t]
    \centering
    \includegraphics[width=0.98\textwidth]
    {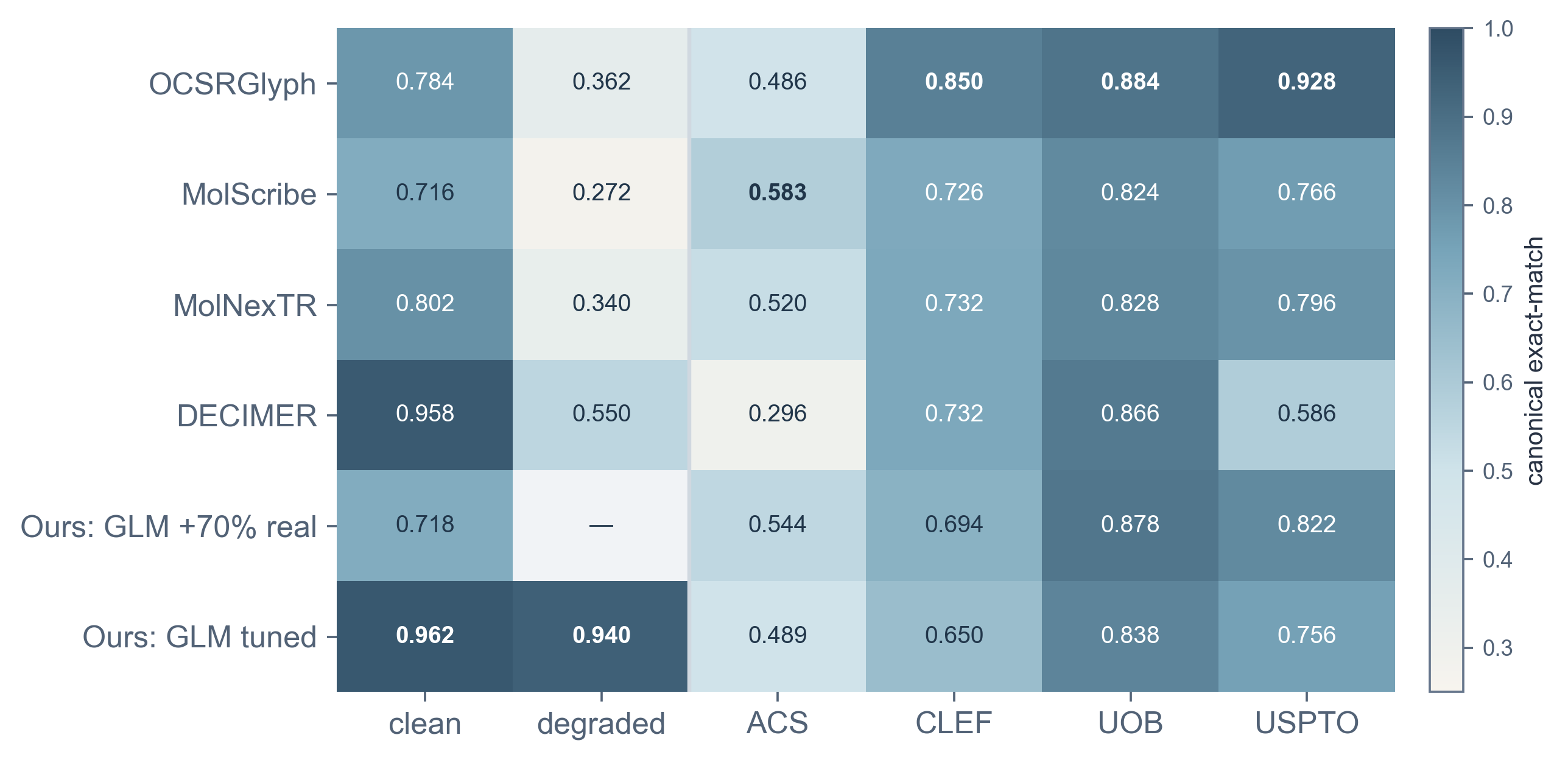}
    \caption{Recognizer rankings change across image conditions, with no system leading on every dataset.}
    \label{fig:leadership}
\end{figure}

Figure~\ref{fig:leadership} presents a representative subset of the comparison.
The resulting ranking is strongly condition-dependent. Specialist recognizers
remain strongest on several real-document sources, whereas the tuned GLM
checkpoint is strongest on the displayed rendered conditions. No recognizer
leads across all six conditions.

\subsection{Specialist OCSR systems}
\label{sec:engines}

Among the released specialist recognizers, OCSRGlyph performs best on the two
patent-derived test sets, reaching $0.850$ on CLEF-IP and $0.928$ on USPTO.
The latter is within one percentage point of the $0.938$ reported on the full
$5{,}719$-image USPTO benchmark. Because the evaluation slice is smaller, this
agreement should be interpreted as a consistency check rather than an exact
reproduction of the published result.

The highest ACS and UOB scores are obtained by the $70\%$-real GLM checkpoint
trained under the full recipe, at $0.598$ and $0.898$. MolScribe follows on ACS
at $0.583$ and OCSRGlyph on UOB at $0.884$; neither margin is significant under
a paired test, as reported below. Leadership on rendered conditions is similarly divided:
the tuned GLM checkpoint leads clean depictions, DECIMER leads CoordGen,
the synthetic-only Qwen checkpoint leads Indigo, and the tuned InternVL
checkpoint leads the degraded condition. Thus, neither a specialist
recognizer nor a fine-tuned VLM dominates the complete evaluation suite.

\begin{table*}[t]
\centering
\small
\caption{All recognizers are compared under the same evaluation and scoring procedure.}
\label{tab:sota}
\setlength{\tabcolsep}{4pt}
\resizebox{\textwidth}{!}{%
\begin{tabular}{lcccccccc}
\toprule
 & \multicolumn{4}{c}{rendered} & \multicolumn{4}{c}{real documents} \\
\cmidrule(lr){2-5}\cmidrule(lr){6-9}
system & clean & CoordGen & Indigo & degraded
       & ACS & CLEF-IP & UOB & USPTO \\
\midrule
\multicolumn{9}{l}{\emph{released specialist recognizers}}\\
\quad OCSRGlyph \citep{glyph}
    & .784 & .714 & .629 & .362 & .486 & .850 & .884 & .928\\
\quad MarkushGlyph \citep{glyph}
    & .554 & .620 & .650 & .488 & .538 & .630 & .878 & .728\\
\quad MolScribe \citep{molscribe}
    & .716 & .700 & .691 & .272 & .583 & .726 & .824 & .766\\
\quad MolNexTR \citep{molnextr}
    & .802 & .868 & .940 & .340 & .520 & .732 & .828 & .796\\
\quad DECIMER \citep{decimer2}
    & .958 & .952 & .957 & .550 & .296 & .732 & .866 & .586\\
\midrule
\multicolumn{9}{l}{\emph{fine-tuned VLM configurations}}\\
\quad GLM-4.1V-9B $+$ $70\%$ real, full recipe
    & .734 & --- & --- & --- & .598 & .774 & .898 & .848\\
\quad GLM-4.1V-9B $+$ $70\%$ real (sweep cell)
    & .718 & .718 & .740 & .656 & .544 & .694 & .878 & .822\\
\quad InternVL3-8B $+$ $70\%$ real (sweep cell)
    & .706 & .674 & .705 & .596 & .538 & .680 & .860 & .770\\
\quad Qwen2.5-VL-7B $+$ $70\%$ real (sweep cell)
    & .680 & .670 & .694 & .552 & .450 & .634 & .854 & .756\\
\quad GLM-4.1V-9B $+$ real, light vision LoRA
    & .726 & .734 & .767 & .688 & .547 & .630 & .876 & .780\\
\quad GLM-4.1V-9B $+$ real, tuned
    & .962 & .948 & .959 & .940 & .489 & .650 & .838 & .756\\
\quad InternVL3-8B $+$ real, tuned
    & .958 & .938 & .954 & .956 & .459 & .610 & .830 & .732\\
\quad Qwen2.5-VL-7B \texttt{sft\_v5}, synthetic only
    & .954 & .950 & .968 & .948 & .169 & .140 & .518 & .116\\
\bottomrule
\end{tabular}}
\end{table*}

The released MarkushGlyph checkpoint provides an additional comparison with a
general-VLM-plus-LoRA design. It is evaluated with the package's no-extension
prompt because the test images contain ordinary single molecules rather than
Markush structures. Its USPTO score of $0.728$ is higher than the $0.690$
reported for the corresponding published evaluation, indicating that the
common harness does not systematically disadvantage this checkpoint.

The controlled $70\%$-real cells also remain competitive despite their fixed,
deliberately untuned $10$k-step recipe: the GLM sweep cell is second only to
the full-recipe checkpoint that shares its mixture, and ahead of both tuned
checkpoints on CLEF-IP, UOB, and USPTO. This comparison is descriptive: the
sweep cells and tuned checkpoints differ in both mixture and training recipe,
so their differences cannot be attributed to tuning alone.

Paired McNemar tests on identical images place the $70\%$-real GLM checkpoint
level with the strongest released system on both journal and hand-drawn
depictions: ACS $0.598$ against MolScribe's $0.583$ ($p{=}0.66$) and UOB
$0.898$ against OCSRGlyph's $0.884$ ($p{=}0.12$). On those two sets it is
significantly ahead of every other released system --- on ACS against MolNexTR
($p{=}0.010$), OCSRGlyph ($p{=}1.6\times10^{-4}$) and MarkushGlyph
($p{=}0.015$), and on UOB against DECIMER ($p{=}8.6\times10^{-4}$) and MolNexTR
($p{=}3.6\times10^{-7}$). Both losses fall on the patent-derived sets and both
are unambiguous: USPTO $0.848$ against $0.928$ ($p{=}4.6\times10^{-7}$) and
CLEF-IP $0.774$ against $0.850$ ($p{=}1.9\times10^{-6}$). The $1.5$- and
$1.4$-point leads on ACS and UOB are therefore reported as parity rather than
as wins.

The ranking further depends on the treatment of stereochemistry. Canonical
exact match requires agreement in molecular connectivity, tetrahedral
chirality, and cis/trans geometry. The chirality-relaxed convention used below
ignores cis/trans geometry, while graph exact match ignores all
stereochemistry. These conventions are reported separately in
Table~\ref{tab:stereo} rather than mixed across systems.

MolScribe illustrates the importance of this distinction. Its USPTO score
increases from $0.766$ under canonical matching to $0.928$ under graph
matching, substantially narrowing the difference from its published graph-level
score of $0.946$. The remaining $0.018$ difference may reflect the smaller
evaluation slice or other evaluation details and should not be attributed
entirely to stereochemistry.

\begin{table}[!htbp]
\centering
\small
\caption{Stereochemistry conventions change reported exact match by up to $18.8$ percentage points.}
\label{tab:stereo}
\setlength{\tabcolsep}{3.5pt}
\resizebox{\columnwidth}{!}{%
\begin{tabular}{lcccccccccccc}
\toprule
 & \multicolumn{3}{c}{ACS}
 & \multicolumn{3}{c}{CLEF-IP}
 & \multicolumn{3}{c}{UOB}
 & \multicolumn{3}{c}{USPTO}\\
\cmidrule(lr){2-4}
\cmidrule(lr){5-7}
\cmidrule(lr){8-10}
\cmidrule(lr){11-13}
system
& can. & chir. & graph
& can. & chir. & graph
& can. & chir. & graph
& can. & chir. & graph\\
\midrule
OCSRGlyph \citep{glyph}
& .486 & .498 & .523
& .850 & .850 & .872
& .884 & .886 & .888
& .928 & .928 & .948\\

MarkushGlyph \citep{glyph}
& .538 & .565 & .619
& .630 & .640 & .818
& .878 & .886 & .888
& .728 & .760 & .832\\

MolScribe \citep{molscribe}
& .583 & .607 & .646
& .726 & .772 & .868
& .824 & .828 & .890
& .766 & .766 & .928\\

MolNexTR \citep{molnextr}
& .520 & .535 & .556
& .732 & .778 & .782
& .828 & .834 & .834
& .796 & .796 & .808\\

DECIMER \citep{decimer2}
& .296 & .314 & .344
& .732 & .758 & .776
& .866 & .870 & .872
& .586 & .598 & .614\\

GLM, $70\%$ real, full recipe
& .598 & .631 & .704
& .774 & .784 & .856
& .898 & .904 & .906
& .848 & .870 & .924\\

GLM, light vision LoRA
& .547 & .583 & .634
& .630 & .644 & .770
& .876 & .882 & .884
& .780 & .802 & .850\\

InternVL3, tuned
& .459 & .498 & .520
& .610 & .620 & .670
& .830 & .838 & .840
& .732 & .742 & .776\\

GLM, vision frozen
& .486 & .517 & .550
& .586 & .598 & .680
& .866 & .872 & .874
& .684 & .700 & .740\\
\bottomrule
\end{tabular}}
\end{table}

The canonical-to-graph difference is system-dependent. On USPTO, the
difference is only $2.0$ points for OCSRGlyph but $16.2$ points for MolScribe
and $10.4$ points for MarkushGlyph. On CLEF-IP, the corresponding MarkushGlyph
difference reaches $18.8$ points. These results are consistent with stronger
stereochemical handling by OCSRGlyph, but they do not isolate whether that
advantage arises from its decoder, training data, or another component of the
system.

For completeness, Table~\ref{tab:allmodels} records additional fine-tuned
checkpoints and the released specialist recognizers evaluated under the common
harness. This inventory combines checkpoints from several experimental series;
percentage labels should therefore not be identified with cells from the
controlled sweep in Section~\ref{sec:realfrac} unless explicitly marked as
sweep cells. Unevaluated or incomplete checkpoints are retained as dashes
rather than interpreted as recognition failures.

\begin{table*}[t]
\centering
\small
\caption{Additional checkpoints and specialist recognizers evaluated using canonical exact match.}
\label{tab:allmodels}
\setlength{\tabcolsep}{4pt}
\resizebox{\textwidth}{!}{%
\begin{tabular}{lccccccccl}
\toprule
 & \multicolumn{4}{c}{rendered}
 & \multicolumn{4}{c}{real documents} & \\
\cmidrule(lr){2-5}\cmidrule(lr){6-9}
configuration
& clean & CoordGen & Indigo & degraded
& ACS & CLEF-IP & UOB & USPTO & note\\
\midrule
\multicolumn{10}{l}{\emph{Qwen2.5-VL}}\\
\quad Qwen2.5-VL-3B, synthetic only
& .752 & .718 & .854 & .670 & .106 & .068 & .298 & .104 & \\

\quad Qwen-7B $+$ real ($\sim\!12\%$, non-grid)
& .786 & .754 & .875 & .732 & .402 & .628 & .842 & .824 & \\

\quad Qwen-7B $+$ real ($\sim\!10\%$, non-grid)
& .940 & .940 & .957 & .916 & .372 & .608 & .818 & .698 & \\

\quad Qwen-7B \texttt{sft\_v6}
& .918 & .904 & .949 & .874 & .347 & .604 & .808 & .742 & \\

\quad Qwen-7B $+$ real ($70\%$, non-grid)
& .696 & .718 & .740 & .614 & .459 & .614 & .832 & .728
& independent run\\

\quad Qwen-7B \texttt{sft\_v5}
& .954 & .950 & .968 & .948 & .169 & .140 & .518 & .116 & \\

\quad Qwen-7B \texttt{sft\_v4}, multi-toolkit $+$ degraded
& .950 & .938 & .957 & .914 & .154 & .124 & .522 & .122
& LM-only LoRA\\

\quad Qwen-7B \texttt{sft\_v3}, single toolkit
& .770 & .742 & .409 & .710 & .112 & .132 & .542 & .098 & \\
\midrule

\multicolumn{10}{l}{\emph{InternVL3}}\\
\quad InternVL3-8B $+$ real, light vision LoRA
& .712 & .694 & .732 & .650 & .508 & .586 & .842 & .744 & \\

\quad InternVL3-8B $+$ real, tuned
& .958 & .938 & .954 & .956 & .459 & .610 & .830 & .732 & \\

\quad InternVL3-8B $+$ real ($20\%$, non-grid)
& .724 & .708 & .821 & .666 & .462 & .568 & .814 & .646 & \\

\quad InternVL3-8B $+$ real ($10\%$, non-grid)
& .802 & .786 & .881 & .760 & .381 & .444 & .798 & .522 & \\

\quad InternVL3-14B, synthetic only
& .896 & .878 & .924 & .880 & .048 & .008 & .114 & .010 & \\

\quad InternVL3-8B, synthetic only
& .914 & .906 & .935 & .874 & .030 & .004 & .070 & .008 & \\
\midrule

\multicolumn{10}{l}{\emph{GLM-4.1V}}\\
\quad GLM-4.1V-9B $+$ $70\%$ real, full recipe
& .734 & --- & --- & --- & .598 & .774 & .898 & .848
& best real-document\\

\quad GLM-4.1V-9B $+$ real, light vision LoRA
& .726 & .734 & .767 & .688 & .547 & .630 & .876 & .780 & \\

\quad GLM-4.1V-9B $+$ real, tuned
& .962 & .948 & .959 & .940 & .489 & .650 & .838 & .756
& balanced point\\

\quad GLM-4.1V-9B $+$ real, vision frozen
& .696 & .676 & .743 & .586 & .486 & .586 & .866 & .684 & \\

\quad GLM-4.1V-9B $+$ real ($10\%$, non-grid)
& .752 & .738 & .862 & .718 & .456 & .466 & .828 & .596 & \\

\quad GLM-4.1V-9B, synthetic only
& .842 & .786 & .897 & .798 & .136 & .054 & .386 & .052 & \\

\quad GLM-4.1V-9B, vision fully unfrozen
& --- & --- & --- & --- & --- & --- & --- & ---
& vision weights not persisted\\
\midrule

\multicolumn{10}{l}{\emph{released specialist recognizers}}\\
\quad OCSRGlyph \citep{glyph}
& .784 & .714 & .629 & .362 & .486 & .850 & .884 & .928
& specialist\\

\quad MarkushGlyph \citep{glyph}
& --- & --- & --- & --- & .538 & .630 & .878 & .728
& general VLM $+$ LoRA\\

\quad DECIMER \citep{decimer2}
& .958 & .952 & .957 & .550 & .296 & .732 & .866 & .586 & \\

\quad MolScribe \citep{molscribe}
& .716 & .700 & .691 & .272 & .583 & .726 & .824 & .766 & \\

\quad MolNexTR \citep{molnextr}
& .802 & .868 & .940 & .340 & .520 & .732 & .828 & .796 & \\
\bottomrule
\end{tabular}}
\end{table*}

\subsection{Published chemical VLMs}
\label{sec:chemvlm}

ChemVLM and TinyChemVL provide an independent comparison with
chemistry-specific multimodal systems. Their pretrained bases, parameter
counts, training corpora, objectives, and inference procedures differ from
those used in the controlled experiments above. The comparison therefore
measures delivered performance under a common evaluation harness rather than
isolating the effect of any single design choice.

All systems are evaluated on the same capped image sets and label files.
Canonical exact match measures complete molecular identity under the strict
stereochemistry convention, while ECFP4 Tanimoto similarity provides a softer
measure of structural agreement. Each external baseline is evaluated using
its selected prompt, tiling, and decoding configuration.

\begin{table}[!htbp]
\centering
\small
\caption{Published chemical VLMs and a generic document VLM evaluated on the same benchmark suite.}
\label{tab:chemvlm}
\setlength{\tabcolsep}{3.5pt}
\resizebox{\textwidth}{!}{%
\begin{tabular}{lcccccccc}
\toprule
model
& clean & CoordGen & Indigo & degraded
& ACS & CLEF-IP & UOB & USPTO\\
\midrule
\multicolumn{9}{l}{\emph{canonical exact match}}\\
generic document VLM
(Qwen2.5-VL-7B, extraction prompt)
& .012 & .008 & .008 & .006
& .030 & .006 & .038 & .004\\

ChemVLM-8B \citep{chemvlm}
& .200 & .228 & .211 & .078
& .311 & .420 & .796 & .554\\

ChemVLM-26B-1.2 \citep{chemvlm}
& .348 & .394 & .298 & .218
& .453 & .608 & .862 & .792\\

TinyChemVL-4B \citep{tinychemvl}
& .678 & .664 & .658 & .404
& .417 & .542 & .888 & .592\\

InternVL3-8B $+$ $\sim\!10\%$ real
& .958 & .938 & .954 & .956
& .459 & .610 & .830 & .732\\
\midrule

\multicolumn{9}{l}{\emph{mean ECFP4 Tanimoto /
fraction with Tanimoto $=1.0$}}\\
generic document VLM
& .23/.02 & .21/.01 & .20/.01 & .18/.01
& .15/.03 & .17/.01 & .27/.04 & .19/.00\\

ChemVLM-8B
& .68/.24 & .70/.29 & .60/.28 & .42/.10
& .55/.35 & .75/.50 & .90/.81 & .83/.58\\

ChemVLM-26B-1.2
& .81/.46 & .85/.58 & .68/.42 & .60/.29
& .65/.52 & .83/.66 & .94/.87 & .93/.82\\

TinyChemVL-4B
& .98/.97 & .97/.95 & .98/.96 & .75/.65
& .63/.48 & .77/.58 & .94/.90 & .82/.61\\

InternVL3-8B $+$ $\sim\!10\%$ real
& 1.00/1.00 & 1.00/1.00 & 1.00/1.00 & .99/.99
& .64/.52 & .80/.67 & .92/.85 & .86/.78\\
\bottomrule
\end{tabular}}
\end{table}

The InternVL checkpoint is substantially stronger than the published chemical
VLMs on all four rendered conditions. Relative to ChemVLM-26B-1.2, canonical
exact match increases by $61.0$ points on clean depictions and $73.6$ points
on the degraded condition. These margins are statistically significant under
the paired tests reported in Appendix~\ref{app:stats}.

The real-document comparison is more mixed. InternVL and ChemVLM-26B-1.2 are
statistically indistinguishable on ACS ($0.459$ versus $0.453$) and CLEF-IP
($0.610$ versus $0.608$). ChemVLM-26B-1.2 is higher on UOB
($0.862$ versus $0.830$) and USPTO ($0.792$ versus $0.732$), with both
differences significant in the paired evaluation. TinyChemVL also exceeds
InternVL on UOB, while InternVL is higher on ACS, CLEF-IP, and USPTO. The
generic document VLM rarely produces the correct molecular structure on
either rendered or real-document images.

The similarity results provide a complementary interpretation. Several
chemistry-specific VLMs produce outputs that are structurally close to the
target even when canonical exact match fails. For example, ChemVLM-26B-1.2
reaches a mean Tanimoto similarity of $0.93$ on USPTO despite a canonical
exact match of $0.792$. Under the ECFP4 implementation used here, a Tanimoto
value of $1.0$ does not guarantee equality under the stricter canonical
stereochemistry convention. Consequently, the Tanimoto@$1.0$ fraction can
exceed canonical exact match and should not be interpreted as an alternative
identity metric.

Across the specialist and chemical-VLM comparisons, leadership remains
condition-specific. Specialist recognizers lead the four real-document
columns: MolScribe on ACS and OCSRGlyph on CLEF-IP, UOB, and USPTO. Fine-tuned
VLM checkpoints lead the rendered conditions, including the constructed
degradation test. The external comparison therefore supports a diagnostic
conclusion rather than universal performance leadership. Representative real
training images provide the most consistent cross-domain improvement in the
controlled experiments, while the preferred base and adaptation surface
remain dependent on the training data composition and fine-tuning strategy.

\FloatBarrier
\section{Cross-Task Probes Beyond Chemistry}
\label{sec:generalize}

The base-model comparison is extended to two small structured-prediction
probes outside chemistry: handwritten image-to-LaTeX recognition
\citep{im2latex} and chart-to-table conversion on a ChartQA-derived test
slice \citep{chartqa}. These experiments are designed as reduced base-swap
comparisons. The pretrained VLM base is varied while the task-specific
training data, fine-tuning recipe, and evaluation procedure are held fixed.

The purpose of these probes is limited. They test whether the base ordering
observed in OCSR also appears on other image-to-structure tasks; they do not
constitute a broad evaluation of general visual reasoning. Each configuration
is represented by one training run, and no confidence intervals are available.

\subsection{Handwritten image-to-LaTeX}

For handwritten mathematical expressions, InternVL3 obtains an edit
similarity of $0.829$ and an exact match of $0.412$. Qwen2.5-VL is close in
edit similarity at $0.815$, although its exact match is lower at $0.354$.
GLM-4.1V obtains an edit similarity of $0.737$. The $0.014$ edit-similarity
difference between InternVL3 and Qwen is small relative to the uncertainty
expected from a single training run, so it should not be interpreted as a
stable ranking without replication. The larger separation from GLM shows,
however, that the GLM advantage observed in parts of the OCSR comparison does
not transfer uniformly to handwritten mathematical recognition.

\subsection{Chart-to-table conversion}

On chart-to-table conversion, InternVL3 obtains an edit similarity of $0.606$,
followed by Qwen2.5-VL at $0.560$ and GLM-4.1V at $0.524$. This ordering agrees
with the handwritten-math probe but differs from the ordering observed in
several real-OCSR regimes. Base-model performance is therefore
task-dependent: a base that transfers well to molecular depictions is not
necessarily the strongest base for other image-to-structure tasks.

\begin{table}[!htbp]
\centering
\small
\caption{VLM base choice also affects performance on non-chemical image-to-sequence tasks.}
\label{tab:cross-task}
\setlength{\tabcolsep}{6pt}
\begin{tabular}{@{}lccc@{}}
\toprule
task & InternVL3 & Qwen2.5-VL & GLM-4.1V\\
\midrule
handwritten image-to-LaTeX
& .829 (.412)
& .815 (.354)
& .737 (---)\\

chart-to-table
& .606 (---)
& .560 (---)
& .524 (---)\\
\bottomrule
\end{tabular}
\end{table}

Taken together, the probes support two narrow conclusions. First, base choice
remains visible outside OCSR, although the small InternVL--Qwen difference on
handwritten mathematics is not sufficient to establish a reliable ranking.
Second, the ordering is task-dependent: GLM ranks last on both cross-task
probes despite performing strongly in several real-OCSR comparisons. The
experiments therefore do not support a task-general notion of visual
robustness associated with any single pretrained base. They also do not
identify whether the observed differences arise from the vision encoder,
connector, language decoder, image processor, or pretraining exposure.

\FloatBarrier
\section{Discussion}
\label{sec:discussion}

OCSR is often treated as a transcription problem that can be improved by
rendering more molecules or adding synthetic corruptions. The results instead
identify document-level depiction shift as the principal difficulty. Real
figures contain renderer conventions, rasterization, compression, cropping,
and annotations that are poorly reproduced by synthetic augmentation. High
SMILES validity but low exact match further indicates that many failures concern
molecular identity rather than output syntax.

Representative real depictions are therefore the most effective intervention
examined here. Real-document accuracy increases with the real-data fraction
across all three bases, although rendered-clean accuracy decreases by
$9.8$--$13.8$ points. Base rankings also change with the real-data regime, and
vision adaptation ranges from ineffective for Qwen2.5-VL-7B to substantial for
InternVL3-8B. Base and adaptation choices should therefore be evaluated near
the real-data fraction intended for deployment.

The main contribution is a controlled separation of real-data fraction,
pretrained base, and adaptation surface. The resulting evidence shows that
representative real supervision provides the most consistent cross-document
gain, while specialist and VLM leadership remains condition-specific. An
evaluation audit additionally shows that model-specific image preprocessing can
artificially enlarge the measured synthetic-to-real gap, making persisted
predictions and real-image control conditions essential.

\section{Limitations}

\paragraph{Base and adaptation scope.}
\label{sec:limits-surface}\label{sec:limits-stack}\label{sec:limits-probe}
Adaptation-surface comparisons are incomplete: the matched Qwen contrast is
limited to $0\%$ real data, and no frozen-vision arm is available at a high real
fraction. The results therefore show that vision adaptation is base- and
data-dependent, but not that it is generally unnecessary. Base comparisons are
also whole-stack comparisons involving different encoders, processors,
tokenizers, decoders, and pretraining corpora; the observed ranking changes
cannot be attributed to a single component.

\paragraph{Evaluation and external comparisons.}
\label{sec:limits-degraded}\label{sec:limits-engines}
\label{sec:limits-external}
The degraded benchmark uses the same corruption generator represented in the
training mixture and therefore measures matched synthetic robustness rather
than transfer to real scans. Comparisons with released specialist and chemical
VLM checkpoints are also not controlled for architecture, training data, or
optimization. Pixel-identical overlap was removed where possible, but exposure
to shared public corpora cannot be excluded. The per-condition best results
come from multiple checkpoints and should be interpreted as a performance
envelope, not a single deployable state-of-the-art system.

\paragraph{Training design and uncertainty.}
\label{sec:limits-sweep}\label{sec:limits-decoding}
Each cell in the $18$-run real-fraction sweep uses one seed and one fixed
recipe. Reported confidence intervals therefore capture evaluation-set
uncertainty, not training-seed variation. The flagship InternVL checkpoint also
uses a different recipe from the nominally similar $10\%$ sweep cell, so it
does not establish a post-$10\%$ plateau. Constrained decoding,
higher-resolution tiling, and sampling-based self-consistency remain untested
and could alter the practical cost--accuracy trade-off.

Image count is also not separated from chemical diversity: all
$365{,}776$ entries in the real pool are distinct molecules with one
image each, so raising the real fraction adds images and molecules
together. The results therefore describe a one-image-per-molecule
regime and do not indicate how many depictions of the same structure
are worth collecting.

\paragraph{Data and task coverage.}
Consensus-derived examples remain weak labels until manually audited. Some
patent results may benefit from molecular overlap with public training sources
and are therefore excluded from headline claims. The study is restricted to
$3$--$14$B bases and primarily single-molecule depictions; Markush structures,
R-groups, reaction diagrams, and severely degraded historical documents remain
out of scope. The two cross-task probes are insufficient to establish
generalization beyond OCSR.

\section{Conclusion}
Real depictions are the most effective lever examined for OCSR under
document-level distribution shift. In the independent Qwen2.5-VL dose series,
ACS exact match rises from $0.154$ without real data to $0.372$ at $9.5\%$
real and $0.459$ at $50.2\%$, while the patent sets improve from approximately
$0.12$ to $0.70$--$0.82$. Although this series is descriptive because its
recipe and training budget also vary, the controlled three-base sweep confirms
that real-document accuracy increases consistently with the real-data
fraction. This gain carries a rendered-domain cost, showing that OCSR training
is an allocation problem between synthetic coverage and deployment-relevant
depictions rather than a simple question of dataset scale. Model choices are
also conditional on this allocation: the between-base spread contracts from
$0.212$ at $0\%$ real data to $0.060$ at $70\%$, and the ranking changes across
the sweep. Vision adaptation similarly produces no measurable ACS benefit for
the matched Qwen experiment ($p{=}1.00$), but improves InternVL3-8B by
$22.8$--$34.6$ points. These results suggest that representative real data
should be acquired first, the pretrained base should be selected at the
intended real-data scale, and the vision path should be adapted only when
supported by a matched comparison. Because labeled real figures remain scarce,
the recognizer is deployed as one voter in a conservative literature-mining
agreement gate, closing a practical loop in which better real-data training
improves recognition and more reliable recognition produces the real-data
resource required by subsequent models.

\section*{Acknowledgments}
We thank DECIMER, MolScribe, and
MolNexTR for open-sourcing their engines. Experiments were run on NVIDIA H200 GPUs.

\bibliographystyle{plainnat}
\bibliography{references}

\appendix

\section{Training and Evaluation Details}
\label{app:train}

\subsection{Fine-tuning configuration}

Unless otherwise specified, LoRA adapters with rank $16$ and
$\alpha=32$ are applied to the attention and MLP projections of the language
decoder. The controlled real-fraction grid additionally uses the same light
vision-LoRA surface for every cell. Language-decoder-only, frozen-vision, and
alternative vision-adaptation configurations are identified explicitly in
the corresponding ablations in Section~\ref{sec:vision}.

Loss is applied only to the assistant SMILES tokens. Prompt tokens, image
tokens, and padding positions are masked with the ignore index $-100$.
Training uses bf16 arithmetic, gradient checkpointing, a learning rate of
$10^{-4}$, cosine decay, and a $3\%$ warm-up period. The per-device batch size
is $8$, with two gradient-accumulation steps.

The controlled $3\times6$ real-fraction experiment is trained for exactly
$10{,}000$ optimization steps per cell rather than for a fixed number of
epochs. At the approximately $496$k-sample mixture size, this corresponds to
roughly $0.32$ epoch under the effective batch configuration. Independent dose
series and tuned checkpoints use the step budgets recorded in
Table~\ref{tab:compute}; they should not be treated as cells of the controlled
sweep.

\subsection{Training data}

Synthetic depictions are generated from a curated one-million-molecule
collection using a leakage-controlled, heavy-atom-stratified split. The
single-renderer condition uses RDKit depictions. The multi-toolkit condition
additionally includes CoordGen, Indigo, and CDK/RanDepict renderings.

After filtering and deduplication, the real-image pool contains approximately
$366$k labeled depictions drawn from USPTO-680k, DECIMER-HDM, MolParser, and an
internal literature-derived collection of approximately $41$k images.
Molecular identities appearing in any evaluation set are excluded using an
InChIKey blocklist containing $55{,}267$ keys. This molecular-identity filter
removes exact identity overlap, although it cannot make every depiction
source equally out of distribution.

\subsection{Evaluation protocol}

The rendered evaluation suite contains clean RDKit depictions, CoordGen
relayouts, Indigo renders, and synthetically degraded images. Each rendered
condition is capped at $500$ images, except Indigo, for which $369$ valid
examples are available. The real-document suite contains ACS ($n=331$),
CLEF-IP ($n=500$), UOB ($n=500$), and USPTO ($n=500$).

Predicted SMILES are parsed and canonicalized with RDKit. Canonical exact
match requires equality under the complete stereochemistry convention used
throughout the main results. Molecular identity is additionally checked
through the corresponding InChIKey. Validity is the proportion of outputs
that can be parsed as molecules by RDKit. All compared systems are evaluated
on the same image files, reference labels, canonicalization implementation,
and scoring code.

\section{Consensus Reconciliation}
\label{app:consensus}

The consensus component is used to construct conservative weak labels from
multiple independently generated OCSR predictions. Its broader design,
calibration, and evaluation are reported separately \citep{verdict}; only the
reconciliation rule needed to interpret the training corpus is summarized
here.

Each engine output is parsed and canonicalized. Salts and solvates are reduced
to the neutral largest fragment, and the resulting parent structure is mapped
to a stereochemistry-preserving InChIKey. Predictions are grouped by this key.
For engines $e=1,\ldots,K$ emitting keys $k_e$, the winning identity and its
agreement count are

\[
k^\star
=
\arg\max_k
\sum_{e=1}^{K}\mathbf{1}\{k_e=k\},
\qquad
a
=
\sum_{e=1}^{K}\mathbf{1}\{k_e=k^\star\}.
\]

Ties are resolved using a fixed engine-priority order. Before acceptance, the
winning structure must pass a substance filter: dummy atoms are rejected and
at least six heavy atoms are required. After this validity and substance
filter, a weak label is emitted only when the agreement count reaches the
specified quorum. All remaining examples are routed to a review queue.

A ground-truth-free round-trip score, obtained by re-rendering a predicted
structure and comparing it with the source crop, is not included in the
acceptance rule because it was not calibrated as a correctness estimator on
real-document figures. The reconciliation layer is deterministic given fixed
engine outputs. Model-specific dependencies are isolated behind uniform
adapters, and the failure of one engine is recorded without terminating the
remaining reconciliation process.

\section{Statistical Methodology}
\label{app:stats}

\subsection{Interpretation of uncertainty intervals}

Each reported checkpoint represents one fine-tuning run evaluated on a fixed
held-out image set. Wilson $95\%$ confidence intervals are calculated for
exact match and validity using the success count obtained directly from the
per-image prediction files. Counts are not reconstructed by rounding an
already reported proportion.

These intervals quantify uncertainty associated with sampling evaluation
images from the corresponding benchmark population. They do not measure
training-seed uncertainty. Multiple-seed training was not performed because
individual runs require between $9.5$ and $37.9$ recorded H200 hours
(Table~\ref{tab:compute}). Consequently, differences between training
configurations should not be interpreted as estimates averaged over
initialization and data-order randomness.

For selected comparisons, two additional analyses are reported. First,
prediction correctness is compared on identical images using an exact
two-sided McNemar test and a paired bootstrap with $10{,}000$ resamples.
Second, selected intermediate checkpoints are reevaluated to measure local
late-training sensitivity. This intermediate-checkpoint comparison does not
replace multi-seed replication and should not be interpreted as a general
bound on training variance.

\subsection{Confidence intervals on real-document exact match}

\begin{table}[!htbp]
\centering
\small
\caption{Canonical exact match and uncertainty on four real-document datasets.}
\label{tab:real-ci}
\resizebox{\textwidth}{!}{%
\begin{tabular}{lcccc}
\toprule
run
& ACS ($n=331$)
& CLEF-IP ($n=500$)
& UOB ($n=500$)
& USPTO ($n=500$)\\
\midrule
InternVL3, synthetic only
& 0.030 {\scriptsize[.016,.055]}
& 0.004 {\scriptsize[.001,.014]}
& 0.070 {\scriptsize[.051,.096]}
& 0.008 {\scriptsize[.003,.020]}\\

InternVL3 $+$ real, tuned
& 0.459 {\scriptsize[.406,.513]}
& 0.610 {\scriptsize[.567,.652]}
& 0.830 {\scriptsize[.795,.860]}
& 0.732 {\scriptsize[.692,.769]}\\

GLM $+$ real, light vision LoRA
& 0.547 {\scriptsize[.493,.600]}
& 0.630 {\scriptsize[.587,.671]}
& 0.876 {\scriptsize[.844,.902]}
& 0.780 {\scriptsize[.742,.814]}\\

GLM $+$ real, vision frozen
& 0.486 {\scriptsize[.433,.540]}
& 0.586 {\scriptsize[.542,.628]}
& 0.866 {\scriptsize[.833,.893]}
& 0.684 {\scriptsize[.642,.723]}\\

Qwen $+$ $\sim10\%$ real, LM-only, non-grid
& 0.060 {\scriptsize[.039,.091]}
& 0.126 {\scriptsize[.100,.158]}
& 0.072 {\scriptsize[.052,.098]}
& 0.172 {\scriptsize[.141,.208]}\\
\bottomrule
\end{tabular}}
\end{table}

Several observed differences are large relative to evaluation-set uncertainty
and the measured late-checkpoint variation. For example, the selected
InternVL configuration increases from $0.030$ to $0.459$ on ACS and from
$0.070$ to $0.830$ on UOB after real training depictions are introduced.
These two checkpoints are not a complete controlled estimate of the
real-fraction effect; the corresponding controlled evidence is provided by
the sweep in Section~\ref{sec:realfrac}.

The matched InternVL vision-surface comparison produces gains of $22.8$ to
$34.6$ percentage points across the four real sets. The matched GLM
vision-surface comparison is smaller, ranging from $1.0$ to $9.6$ points.
For GLM, the nominal McNemar test is significant on ACS, CLEF-IP, and USPTO,
but the CLEF-IP result does not survive a Bonferroni correction over four real
sets. The $0.060$ between-base spread at $70\%$ real data is also sufficiently
small that it should not be interpreted as a stable ranking without
multi-seed replication.

\subsection{Paired comparisons}

For each comparison, let $b$ denote the number of images for which system
$A$ is correct and system $B$ is incorrect, and let $c$ denote the reverse.
Table~\ref{tab:paired-tests} reports the exact two-sided McNemar test and the
paired exact-match difference. Bootstrap confidence intervals are shown where
available. The reported $p$-values are nominal unless an adjustment is stated
explicitly.

\begin{table}[!htbp]
\centering
\small
\caption{Paired comparisons on identical images. $\Delta$ is exact match for
system $A$ minus exact match for system $B$. Bracketed values are paired
bootstrap $95\%$ intervals where available.}
\label{tab:paired-tests}

{
\setlength{\tabcolsep}{5pt}
\renewcommand{\arraystretch}{1.05}

\begin{tabular}{@{}lccccc@{}}
\toprule
set & acc.\ $A$ & acc.\ $B$ & $\Delta$ & $b/c$ & McNemar $p$\\
\midrule

\multicolumn{6}{@{}l}{
\emph{InternVL3: tuned $+$ real ($A$) vs.\ synthetic only ($B$)}}\\
ACS
& .459 & .030
& $+.429$ {\scriptsize[+.375,+.483]}
& 144/2 & $<10^{-3}$\\
CLEF-IP
& .610 & .004
& $+.606$ {\scriptsize[+.564,+.648]}
& 303/0 & $<10^{-3}$\\
UOB
& .830 & .070
& $+.760$ {\scriptsize[+.722,+.798]}
& 380/0 & $<10^{-3}$\\
USPTO
& .732 & .008
& $+.724$ {\scriptsize[+.684,+.762]}
& 362/0 & $<10^{-3}$\\

\midrule
\multicolumn{6}{@{}l}{
\emph{GLM $+$ real: light vision LoRA ($A$) vs.\ frozen vision ($B$)}}\\
ACS
& .547 & .486
& $+.060$ {\scriptsize[+.024,+.100]}
& 31/11 & .003\\
CLEF-IP
& .630 & .586
& $+.044$ {\scriptsize[+.008,+.080]}
& 53/31 & .021\\
UOB
& .876 & .866
& $+.010$ {\scriptsize[-.006,+.026]}
& 11/6 & .332\\
USPTO
& .780 & .684
& $+.096$ {\scriptsize[+.058,+.134]}
& 72/24 & $<10^{-3}$\\

\midrule
\multicolumn{6}{@{}l}{
\emph{InternVL $+$ real at step $9{,}000$: light vision LoRA ($A$) vs.\
frozen vision ($B$)}}\\
ACS
& .505 & .227
& $+.278$
& 100/8 & $2\!\times\!10^{-21}$\\
CLEF-IP
& .578 & .240
& $+.338$
& 176/7 & $2\!\times\!10^{-43}$\\
UOB
& .844 & .616
& $+.228$
& 117/3 & $4\!\times\!10^{-31}$\\
USPTO
& .726 & .380
& $+.346$
& 191/18 & $1\!\times\!10^{-37}$\\

\midrule
\multicolumn{6}{@{}l}{
\emph{External: InternVL3 $+$ real ($A$) vs.\ ChemVLM-26B-1.2 ($B$)}}\\
clean
& .958 & .348
& $+.610$ {\scriptsize[+.564,+.654]}
& 309/4 & $<10^{-3}$\\
degraded
& .954 & .218
& $+.736$ {\scriptsize[+.694,+.776]}
& 372/4 & $<10^{-3}$\\
ACS
& .459 & .453
& $+.006$ {\scriptsize[-.045,+.057]}
& 39/37 & .909\\
CLEF-IP
& .610 & .608
& $+.002$ {\scriptsize[-.046,+.050]}
& 76/75 & 1.000\\
UOB
& .830 & .862
& $-.032$ {\scriptsize[-.060,-.004]}
& 19/35 & .040\\
USPTO
& .732 & .792
& $-.060$ {\scriptsize[-.106,-.016]}
& 52/82 & .012\\
\bottomrule
\end{tabular}
}
\end{table}

The external comparison confirms the condition-specific ranking described in
Section~\ref{sec:chemvlm}. InternVL has large advantages over ChemVLM-26B-1.2
on clean and degraded depictions. ACS and CLEF-IP are statistical ties under
the paired test. ChemVLM-26B-1.2 is higher on UOB and USPTO under the nominal
tests, although correction for multiple comparisons should be considered when
interpreting these four real-set comparisons.

Against ChemVLM-8B, the InternVL checkpoint is higher on ACS, CLEF-IP, and
USPTO ($p\leq10^{-5}$) and is statistically tied on UOB ($p=.086$). Relative
to TinyChemVL-4B, InternVL is higher on all four rendered conditions, CLEF-IP,
and USPTO; ACS is a statistical tie, while TinyChemVL is higher on UOB.
Relative to the generic document-extraction VLM, the InternVL checkpoint is
higher on every rendered and real-document condition.

\subsection{Late-checkpoint sensitivity}

Selected intermediate checkpoints are evaluated to determine whether the
reported final values are unusually sensitive to the last part of training.
The InternVL checkpoint is taken $2{,}000$ steps before the final checkpoint;
the GLM checkpoints are taken $1{,}000$ steps before the final checkpoint.

\begin{table}[!htbp]
\centering
\small
\caption{Final-checkpoint exact match minus intermediate-checkpoint exact
match. }
\label{tab:stability}
\begin{tabular}{lcccc}
\toprule
run & ACS & CLEF-IP & UOB & USPTO\\
\midrule
\texttt{internvl3\_real}
& $-.006$ & $+.026$ & $+.006$ & $-.002$\\
\texttt{glm\_vr1}
& $+.000$ & $-.010$ & $-.002$ & $+.004$\\
\texttt{glm\_vision\_frozen}
& $+.000$ & $+.006$ & $+.002$ & $+.000$\\
\bottomrule
\end{tabular}
\end{table}

All observed final-minus-intermediate differences have magnitude at most
$0.026$. This indicates that the selected results are not determined solely
by the final checkpoint, but it does not establish stability across random
seeds or independently reconstructed training mixtures.

The planned statistical artifact bundle contains the per-image predictions,
paired-test implementation, bootstrap implementation, and the generated
summary files, including \texttt{paired\_tests.csv} and
\texttt{stability.csv}.

\section{Reproducibility}
\label{app:repro}

The retained experiment artifacts include checkpoints for
\texttt{sft\_v4}, \texttt{sft\_real}, \texttt{glm\_vr1},
\texttt{internvl3\_vr1}, \texttt{sft\_qwen\_vr1}, and the vision-surface
ablations. The internal artifact store also contains the real-pool build,
evaluation predictions, run manifests, and the literature-mining pipeline
that maps document figures to molecular crops, engine predictions, and weak
labels.

Evaluation is performed through \texttt{eval\_adapter.py} and
\texttt{eval\_vlm.py}. Model-specific processors and dependency versions are
recorded in the corresponding run manifest. Fitted constants and selection
rules are frozen using training or development data before final test-set
evaluation.

The planned public release includes the evaluation and consensus code, the
real-pool construction recipe, per-image prediction files, and the
$6{,}146$-label weakly supervised corpus released alongside
\citet{verdict}. The associated filtering and evaluation-set deduplication
reduce $9{,}074$ quorum-reaching depictions to $6{,}146$ released labels.
Any private source images that cannot be redistributed are represented by
reproducible identifiers and construction metadata rather than copied into
the public artifact.

\subsection{Compute accounting}

Useful training compute is taken from the Hugging Face
\texttt{total\_flos} field. Recorded elapsed time is taken from the training
loop logs. Across approximately $25$ SFT, real-fraction, vision-ablation, and
preference-training runs, the summed useful compute is approximately
$1.1\times10^{20}$ FLOP, or $110$ EFLOP. The measured elapsed time over runs
with surviving timing logs is approximately $318$ hours. Including untimed
early runs gives an estimated total of approximately $370$--$410$ hours.

Table~\ref{tab:compute} reports the largest individual runs. These figures are
reported separately because useful FLOP and occupied accelerator time measure
different aspects of computational cost. Model-FLOP utilization is not
reported because it requires a verified accelerator world size and a
consistent definition of useful FLOP across all archived runs.

\begin{table}[!htbp]
\centering
\small
\caption{Recorded compute for the largest fine-tuning runs.}
\label{tab:compute}
\begin{tabular}{lrrr}
\toprule
run & steps & useful compute (PFLOP) & wall-clock (h)\\
\midrule
\texttt{internvl3\_vr1}
& 10{,}000 & 11{,}741 & 22.9\\
\texttt{sft\_v5\_morereal}
& 20{,}000 & 9{,}380 & 29.8\\
\texttt{internvl3\_real}
& 20{,}000 & 9{,}240 & 32.3\\
\texttt{sft\_v6}
& 20{,}000 & 6{,}698 & 22.7\\
\texttt{glm\_real\_sftreal}
& 18{,}000 & 6{,}408 & 37.2\\
\texttt{sft\_real}
& 20{,}000 & 6{,}366 & ---\\
\texttt{iv3\_real\_40}
& 6{,}000 & 6{,}088 & 14.9\\
\texttt{sft\_qwen\_vr1}
& 10{,}000 & 4{,}460 & 9.5\\
\texttt{glm\_vr1}
& 10{,}000 & 3{,}653 & 37.9\\
\texttt{glm\_real\_10}
& 10{,}000 & 3{,}237 & 10.5\\
\midrule
approximately $25$ runs
& --- & $\approx110{,}000$ & $\approx370$--$410$\\
\bottomrule
\end{tabular}
\end{table}

\end{document}